\documentclass{article}

\usepackage{verbatim}
\usepackage[utf8]{inputenc} % if you're on pdfLaTeX and your file is UTF-8
\usepackage[T1]{fontenc}
\DeclareUnicodeCharacter{2223}{\ensuremath{\mid}}    % ∣
\DeclareUnicodeCharacter{03B8}{\ensuremath{\theta}}
\usepackage{bm} 
\usepackage{PRIMEarxiv}
\usepackage{amsmath, amssymb}
\usepackage[utf8]{inputenc} % allow utf-8 input
\usepackage[T1]{fontenc}    % use 8-bit T1 fonts
\usepackage{color}
\usepackage{hyperref}       % hyperlinks
\usepackage{url}            % simple URL typesetting
\usepackage{booktabs}       % professional-quality tables
\usepackage{amsfonts}       % blackboard math symbols
\usepackage{nicefrac}       % compact symbols for 1/2, etc.
\usepackage{microtype}      % microtypography
\usepackage{lipsum}
\usepackage{fancyhdr}       % header
\usepackage{graphicx}       % graphics
\usepackage{geometry} % Required for changing page layout
\usepackage{tikz}
\usepackage{pdflscape} % For landscape pages
\usepackage{lipsum} % For generating dummy text
\usetikzlibrary{positioning, arrows.meta, shapes, fit, backgrounds}
 
\tikzset{
    arrow/.style={thick,->,>=Stealth},
    every node/.style={rounded corners=2pt, align=center, font=\small, minimum width=1.7cm, minimum height=0.75cm, text width=1.5cm, text centered}
}

\graphicspath{{media/}}     % organize your images and other figures under media/ folder
\newcommand{\mycomment}[1]{}
\title{From Classical Probabilistic Latent Variable Models to Modern Generative AI: A Unified Perspective}

\author{
  Tianhua Chen \\
  School of Computing and Engineering\\
  University of Huddersfield, UK \\
  \texttt{T.Chen@hud.ac.uk} \\
}

\begin{document}
\maketitle

\begin{abstract}

From large language models to multi-modal agents, Generative Artificial Intelligence (AI) now underpins state-of-the-art systems. Despite their varied architectures, many share a common foundation in probabilistic latent variable models (PLVMs), where hidden variables explain observed data for density estimation, latent reasoning, and structured inference.
This paper presents a unified perspective by framing both classical and modern generative methods within the PLVM paradigm.
We trace the progression from classical flat models such as probabilistic PCA, Gaussian mixture models, latent class analysis, item response theory, and latent Dirichlet allocation, through their sequential extensions including Hidden Markov Models, Gaussian HMMs, and Linear Dynamical Systems, to contemporary deep architectures: Variational Autoencoders as Deep PLVMs, Normalizing Flows as Tractable PLVMs, Diffusion Models as Sequential PLVMs, Autoregressive Models as Explicit Generative Models, and Generative Adversarial Networks as Implicit PLVMs.
Viewing these architectures under a common probabilistic taxonomy reveals shared principles, distinct inference strategies, and the representational trade-offs that shape their strengths. We offer a conceptual roadmap that consolidates generative AI’s theoretical foundations, clarifies methodological lineages, and guides future innovation by grounding emerging architectures in their probabilistic heritage.

\end{abstract}

% keywords can be removed
%\keywords{Probabilistic Latent Variable Models \and Generative AI \and Deep Generative Models \and Variational Inference \and Posterior Inference \and Latent Representation}


\section{Introduction}

Artificial Intelligence (AI) is undergoing a profound shift with generative modeling emerging as a defining paradigm. Just as the introduction of layered architectures catalyzed the deep learning revolution, generative models now represent a higher level of abstraction—one that is rapidly reshaping the foundations of modern AI. This shift has been described as a progression from layers enabling deep networks, to deep networks powering transformative applications, and now to generative models driving systems such as large language models (LLMs), multi-modal agents, and scientific discovery engines~\cite{he2025deeplearningday,jo2023promise}. Generative AI is no longer just an application domain; it is a computational paradigm for modeling, synthesizing, and reasoning about the world~\cite{epstein2023art}.  

At its core, generative modeling concerns learning conditional distributions $P(x|y)$~\cite{salakhutdinov2015learning}, a perspective that unifies a broad range of tasks~\cite{he2025deeplearningday,cao2023comprehensive}. In conversational AI, $y$ may be a text prompt and $x$ the generated reply; in text-to-image systems, $y$ is a description and $x$ the synthesized visual content. Even classical supervised learning can be framed generatively: given an image $y$, we model $P(x|y)$ where $x$ is a class label, caption, or explanation. This formulation naturally extends to open-vocabulary recognition, multi-modal generation, and other tasks where $x$ and $y$ may mix text, images, video, and structured data.  

While generative models vary in architecture and training objectives, many are deeply rooted in the principles of probabilistic latent variable models (PLVMs)—a class of models that introduce hidden variables to explain observed data, enable density estimation, and support structured inference~\cite{bishop2006,blei2003latent}. Classical examples such as Probabilistic principle component analysis, Gaussian mixture models and latent Dirichlet allocation established core ideas: mapping high-dimensional observations into lower-dimensional latent spaces, modeling complex distributions via simpler latent priors, and performing inference over unobserved structure. These ideas remain central to representation learning, data synthesis, and clustering in contemporary systems~\cite{berahmand2024autoencoders}.  

The scope of PLVMs extends naturally from flat models with i.i.d.\ assumptions, to sequential models such as Hidden Markov Models, Gaussian HMMs, and Linear Dynamical Systems, which capture temporal dependencies via latent state dynamics.
%Contemporary generative models build upon and extend these principles. Despite their apparent differences, all can be situated within—or related to—the PLVM framework.   
Contemporary deep generative architectures can likewise be positioned within this probabilistic lineage: Variational Autoencoders (VAEs) as Deep PLVMs ~\cite{kingma2022autoencodingvariationalbayes}, where neural parameterizations expand the expressive power of classical formulations; Normalizing Flows~\cite{kobyzev2020normalizing} as Tractable PLVMs, enabling exact density evaluation and inference via invertible transformations; Diffusion Models~\cite{lipman2022flow,sohl2015deep} as Sequential PLVMs, generating data through stepwise latent evolution; Autoregressive Models as Explicit Probabilistic Generative Models, defining fully factorized likelihoods; and Generative Adversarial Networks (GANs)~\cite{goodfellow2014generative} in turn, as Implicit PLVMs trained via adversarial feedback, bypassing explicit likelihoods altogether. 

The primary contribution of this paper is to provide a unified probabilistic framework that connects classical PLVMs with the latest wave of deep generative architectures. By systematically tracing how foundational ideas are inherited, adapted, or bypassed, we clarify the shared principles, distinctive mechanisms, and future trajectories of generative modeling. This synthesis serves both as a theoretical scaffold for researchers seeking to contextualize modern architectures and as a practical reference for those aiming to develop the next generation of generative models.  

In the sections that follow, we formalize key probabilistic ideas, explore inference strategies and trace how these methods evolve into the powerful tools that define modern generative modelling. The next section offers an overview and outlines the roadmap of this paper, previewing how we unify classical and modern approaches through a decision-tree taxonomy based on posterior dependence, inference tractability, and learning strategies.

\section{Overview and Roadmap}\label{sec:overview}

Building on the unifying perspective outlined in the introduction, we now situate classical and modern generative models within a shared probabilistic latent variable model (PLVM) taxonomy. This section organises the landscape into complementary perspectives—generative structure, inference tractability, and broader modelling dimensions, before presenting a decision-tree roadmap (Figure~\ref{fig:decision_tree_landscape_vertical}) that will guide the remainder of the paper.

\subsection{Latent Variable Models by Generative Structure}
A natural starting point is to classify PLVMs by whether the latent and observed variables are continuous or discrete. This yields four canonical types, each with characteristic assumptions and inference challenges:

\begin{itemize}%[leftmargin=*]
    \item \textbf{Continuous latent--continuous observation:} This category includes models such as Probabilistic Principal Component Analysis (PPCA)~\cite{bishop2006}, Variational Autoencoders (VAE)~\cite{kingma2022autoencodingvariationalbayes}, and Kalman Filters (KF)~\cite{khodarahmi2023review}, which assume that both latent and observed variables are real-valued.

    \item \textbf{Continuous latent--discrete observation:} The Item Response Theory (IRT)~\cite{hambleton2013item} is a representative model here, capturing scenarios where underlying continuous latent variables give rise to categorical outcomes.

    \item \textbf{Discrete latent--discrete observation:} Hidden Markov Models (HMMs)~\cite{bouguila2022hidden} and Latent Class Models (LCMs)~\cite{weller2020latent} are classic models in this category, commonly used in applications involving classification or topic modeling.

    \item \textbf{Discrete latent--continuous observation:} This class includes Gaussian Mixture Models (GMMs)~\cite{bishop2006} and Gaussian-HMMs~\cite{bouguila2022hidden}, modelling data from mixtures of continuous distributions conditioned on discrete latent states.
\end{itemize}

These structural distinctions not only shape the semantics of the generative process but also the choice of inference methods and learning algorithms. They provide the foundational lens through which we begin our discussion. In particular, we introduce a representative model from each of the four types as initial case studies—each exemplifying a different generative structure and offering a platform to develop and compare core probabilistic modeling concepts. %Additional models and variants will be introduced later from complementary perspectives such as inference tractability and optimization techniques.

We begin in Section~\ref{sec:ppca} with PPCA, which represents the continuous latent–continuous observation setting. As a probabilistic extension of classical PCA, PPCA provides an analytically tractable example that introduces key ideas such as latent space representation, posterior inference, and generative sampling. These concepts serve as a conceptual foundation for the models that follow.

Section~\ref{sec:finite} then presents three classic models covering the remaining generative structures: GMMs for discrete latent–continuous observations, LCA for discrete latent–discrete observations, and IRT for continuous latent–discrete observations. Together, these models illustrate the breadth of latent variable modeling scenarios and motivate the need for more general inference tools. 

\subsection{Inference Tractability and Learning Strategies}

The generative structure of a model strongly influences how learning and inference are carried out. 
Models like PPCA admit exact closed-form parameter and posterior solutions. Others, including GMMs and LCA, maintain tractable posteriors but require iterative procedures like Expectation–Maximization (EM)~\cite{bishop2006}, a classical method for maximizing the likelihood when posterior computations are straightforward.  
As discussed in Section~\ref{sec:elbo_em}, the Evidence Lower Bound (ELBO) provides a general framework in these cases, offering a tractable objective that EM can optimize when closed-form updates exist~\cite{bishop2006,ng2022cs229}.

In more complex cases (e.g., LDA, VAEs), posteriors are intractable and must be approximated. 
In these cases, Variational Inference (VI) is often used as an approximate inference strategy. VI reframes posterior estimation as an optimization problem, typically minimizing the Kullback-Leibler (KL) divergence between the true posterior and a chosen family of variational distributions~\cite{goodfellow2016deep}. The development and popularization of VI have played a pivotal role in the success of probabilistic models. We delve deeper into this approach in Section~\ref{sec:vi}, discuss its application to Latent Dirichlet Allocation (LDA)~\cite{baker2004irt}, and further illustrate its use in VAEs in Section~\ref{sec:vae}.

\subsection{Broader Modeling Dimensions: Bayesian, Hierarchical, and Temporal Extensions}

Beyond generative structure and inference tractability, probabilistic latent models may be further understood through several complementary perspectives.
One perspective concerns the underlying inference paradigm. Many classical and modern PLVMs, such as PPCA, GMMs, and VAEs, adopt maximum likelihood estimation for learning model parameters. In contrast, models like LDA and hierarchical Bayesian models employ Bayesian generative modelling, where prior distributions are placed over latent variables and parameters. Bayesian approaches offer principled mechanisms for uncertainty quantification and regularization, particularly valuable in low-data regimes or applications requiring interpretability.

Another perspective lies in the generative hierarchy of the models. Classical models typically operate with a two-layer structure, in which latent variables directly generate observed data. More complex models, such as LDA, introduce an additional layer of latent variables, resulting in deeper three-level hierarchies. Deep latent models like VAEs extend this principle by embedding neural networks into the generative and inference processes, while diffusion models and normalizing flows adapt it in distinctive ways—diffusion models unfolding a sequence of latent states through noising and denoising steps, and flows composing multiple invertible transformations. In each case, the use of neural architectures or multi-stage transformations implicitly defines highly flexible, multi-layered latent structures capable of capturing complex, nonlinear dependencies in data.

A further dimension involves temporal dependencies. While foundational PLVMs often assume i.i.d.\ data, time-series variants replace this assumption with latent Markov processes. Hidden Markov Models (HMMs) and Kalman Filters (KFs) exemplify this setting combining latent state transitions and emission models. As outlined briefly in Section~\ref{sec:time_series}, despite introducing temporal dependencies, both classes admit exact inference through dynamic programming methods such as Forward–Backward or Kalman filtering/smoothing, and can be trained efficiently via the EM algorithm. These principles can further extend to modern sequential models—such as RNN-based VAEs and certain diffusion processes—which generalise the classical frameworks by replacing simple transition/emission mappings with neural networks and using approximate (e.g., variational or amortised) inference in place of exact methods.

\subsection{Probabilistic Latent View of Modern Generative Models}
\label{sec:nonposterior_models}

While many classic PLVMs rely directly on posterior inference to connect latent and observed variables, modern generative models vary in how closely they retain this structure. Variational Autoencoders (VAEs) and Normalizing Flows (NFs) preserve an explicit latent-variable formulation and permit full probabilistic interpretation: VAEs combine an encoder–decoder architecture with variational inference to enable scalable likelihood-based learning, whereas NFs replace approximate inference with a sequence of invertible transformations, allowing exact computation of likelihoods and posteriors. Diffusion Models extend the latent framework to a sequential setting, generating data by reversing a fixed noising process through learned denoising or score-matching objectives, thereby creating a structured latent trajectory without requiring a conventional posterior. Autoregressive Models depart further, factorizing the joint distribution into fully observable conditional probabilities; although they lack explicit latent variables, their ordered structure can mimic latent trajectories and supports exact maximum likelihood estimation. At the other end of the spectrum, Generative Adversarial Networks (GANs) define implicit latent variable models in which a generator maps noise from a simple prior to realistic samples via an adversarial game against a discriminator, bypassing explicit likelihoods altogether.

In this view, contemporary generative models form a continuum from deep latent-variable architectures with exact inference to fully observable or implicit formulations. Even those that abandon classical posterior-based inference can be interpreted as extensions or reinterpretations of PLVM principles, modifying how latent structure is specified, inferred, or circumvented, while striving for expressive and computationally feasible generative processes.

\subsection{A Roadmap}

The perspectives above motivate a unified organisational view of generative models, summarised in Figure~\ref{fig:decision_tree_landscape_vertical}. This figure serves both as a navigation tool for the paper and as a conceptual contribution in its own right, consolidating classical and modern approaches within the probabilistic latent variable model (PLVM) framework. The roadmap classifies models according to three key axes: their reliance on posterior inference, the tractability of that inference, and the corresponding learning strategy.

The decision tree first distinguishes models that explicitly require posterior inference from those that do not. This top-level split reflects a fundamental design choice: whether generation proceeds via latent-variable reasoning or through alternative mechanisms such as adversarial objectives or autoregressive conditioning.

In the posterior-based branch, models are grouped by inference tractability. Some, like PPCA, permit exact closed-form solutions for both parameters and posteriors. Others, such as GMMs and LCA, retain tractable posteriors but rely on iterative procedures like Expectation–Maximization (EM). More complex cases—LDA, IRT, and VAEs—have intractable posteriors and employ variational inference or sampling-based approximations. Specialised families such as Normalizing Flows achieve exact inference through invertible transformations, while Diffusion Models, often framed via ELBO optimisation, can also be trained through score-based methods that approximate posterior means.

The non-posterior branch groups models that bypass explicit posterior computation. GANs operate as implicit latent-variable models trained through adversarial games, while autoregressive models define fully observable processes by factorising the joint distribution into conditionals. Although they differ structurally, both retain core PLVM principles, such as mapping simple priors to complex outputs or modelling dependencies sequentially.

Colour-coding in Figure~\ref{fig:decision_tree_landscape_vertical} links each model class to its primary inference or learning paradigm, visually mapping conceptual categories to methodological properties. While the tree offers a clean classification, many models may straddle multiple categories in practice—for example, PPCA can be trained either in closed form or via EM for improved numerical stability.  

The sections that follow trace this taxonomy from classical models to modern deep architectures, with mathmatical notation introduced locally for clarity. In doing so, the roadmap situates diverse generative models within a single probabilistic perspective, providing a consistent frame for both historical context and contemporary developments.

\begin{landscape}
\begin{figure}[t]
    \centering
    \begin{tikzpicture}[
        scale=0.7,
        transform shape,
        node distance=1.2cm and 2.0cm,
        yshift=1.0cm
    ]
    % Root
    \node (q1) [draw, fill=gray!20] {Q1:\\ Uses Posterior?};

    % Separate boxes for GANs and Autoregressive
    \node (ganbox) [draw, fill=yellow!20, below left=of q1, xshift=-3.6cm, yshift=-0.5cm, minimum width=5.2cm, minimum height=2.0cm, align=left] {
        \textbf{GANs}\\
        $\cdot$Latent variables\\
        $\cdot$No likelihood\\
        $\cdot$Adversarial training\\
        (Section~\ref{sec:GANs})
    };

    \node (arbox) [draw, fill=yellow!10, below=of ganbox, yshift=-0.3cm, minimum width=5.2cm, minimum height=2.0cm, align=left] {
        \textbf{Autoregressive Models}\\
        $\cdot$No latent variables\\
        $\cdot$Fully observable\\
        $\cdot$Factorized likelihood\\
        (Section~\ref{sec:autoregressive})
    };

    % Posterior-based branching
    \node (q2) [draw, fill=gray!20, below right=of q1, xshift=0.7cm, yshift=-0.5cm] {Q2:\\ Closed-form Solution?};
    \node (ppca) [draw, fill=blue!10, below left=of q2, xshift=-0.7cm, yshift=-0.5cm] {PPCA\\(Section~\ref{sec:ppca})};
    \node (q3) [draw, fill=gray!20, below right=of q2, xshift=0.7cm, yshift=-0.5cm] {Q3:\\ Posterior tractable?};

    \node (q4) [draw, fill=gray!20, below left=of q3, xshift=-0.7cm, yshift=-0.5cm] {Q4:\\ Uses EM?};
    \node (q5) [draw, fill=gray!20, below right=of q3, xshift=0.7cm, yshift=-0.5cm] {Q5:\\ Posterior Approximation?};

    \node (q6) [draw, fill=gray!20, below left=of q4, xshift=-0.6cm, yshift=-0.5cm] {Q6:\\ i.i.d. or Sequential Data?};
    \node (iid) [draw, fill=green!10, below left=of q6, xshift=-0.5cm, yshift=-0.5cm] {GMM, LCA\\(Section~\ref{sec:finite})};
    \node (sequential) [draw, fill=green!10, below right=of q6, xshift=0.4cm, yshift=-0.5cm] {HMM, Kalman Filter\\(Section~\ref{sec:time_series})};

    \node (nf) [draw, fill=cyan!20, below right=of q4, xshift=0.3cm, yshift=-0.5cm] {Normalizing Flows\\(Section~\ref{sec:nf})};
    \node (vi) [draw, fill=orange!20, below left=of q5, xshift=1.2cm, yshift=-0.5cm] {LDA\\(Section~\ref{sec:vi})\\VAE\\(Section~\ref{sec:vae})};
    \node (sampling) [draw, fill=purple!20, below right=of q5, xshift=0.4cm, yshift=-0.5cm] {IRT\\(Section~\ref{sec:finite})};
    \node (scorebased) [draw, fill=teal!20, below=of q5, yshift=-1.0cm] {Diffusion Models\\(Section~\ref{sec:diffusion})};

    % Arrows
    \draw[arrow] (q1) -- ++(-2.4,-1.2) node[midway, left, font=\tiny] {No} -- (ganbox);
 %   \draw[arrow] (q1) -- ++(-2.4,-3.0) -- (arbox);
    \draw[arrow] (q1) -- node[midway, right, font=\tiny] {Yes} (q2);

    \draw[arrow] (q2) -- node[midway, above left, font=\tiny] {Yes} (ppca);
    \draw[arrow] (q2) -- node[midway, above right, font=\tiny] {No} (q3);

    \draw[arrow] (q3) -- node[midway, above left, font=\tiny] {Yes} (q4);
    \draw[arrow] (q3) -- node[midway, above right, font=\tiny] {No} (q5);

    \draw[arrow] (q4) -- node[midway, above left, font=\tiny] {Yes} (q6);
    \draw[arrow] (q4) -- node[midway, above right, font=\tiny] {No} (nf);

    \draw[arrow] (q6) -- node[midway, above left, font=\tiny] {i.i.d.} (iid);
    \draw[arrow] (q6) -- node[midway, above right, font=\tiny] {Sequential} (sequential);

    \draw[arrow] (q5) -- node[midway, above left, font=\tiny] {VI + KL} (vi);
    \draw[arrow] (q5) -- node[midway, above right, font=\tiny] {Sampling} (sampling);
    \draw[arrow] (q5) -- node[midway, right, font=\tiny] {Score-based} (scorebased);

    % Background boxes
    \begin{pgfonlayer}{background}
        \node[draw, dashed, fit=(q2) (q3) (q4) (q5) (q6) (ppca) (iid) (nf) (sequential) (vi) (sampling) (scorebased), label=below:{Posterior-based Models}] {};
        \node[draw, dashed, fit=(ganbox) (arbox), label=below:{Models Without Explicit Posteriors}] {};
    \end{pgfonlayer}

    % Legend
    \matrix [draw, right=5cm of q1, yshift=3.2cm, column sep=0.1cm, row sep=0.05cm, font=\scriptsize\sffamily] (legend) {
        \node[fill=blue!10, draw, minimum size=0.15cm] {}; & \node[font=\scriptsize\sffamily, text width=2.2cm] {Exact Closed-form Inference}; \\
        \node[fill=green!10, draw, minimum size=0.15cm] {}; & \node[font=\scriptsize\sffamily, text width=2.2cm] {Tractable Posterior, EM Applicable}; \\
        \node[fill=cyan!20, draw, minimum size=0.15cm] {}; & \node[font=\scriptsize\sffamily, text width=2.2cm] {Exact Posterior via Invertible Transformations}; \\
        \node[fill=orange!20, draw, minimum size=0.15cm] {}; & \node[font=\scriptsize\sffamily, text width=2.2cm] {Variational Inference + KL}; \\
        \node[fill=purple!20, draw, minimum size=0.15cm] {}; & \node[font=\scriptsize\sffamily, text width=2.2cm] {Sampling-based Inference}; \\
        \node[fill=teal!20, draw, minimum size=0.15cm] {}; & \node[font=\scriptsize\sffamily, text width=2.2cm] {Score-based Inference}; \\
        \node[fill=yellow!20, draw, minimum size=0.15cm] {}; & \node[font=\scriptsize\sffamily, text width=2.2cm] {Implicit Generative Models (no likelihood)}; \\
        \node[fill=yellow!10, draw, minimum size=0.15cm] {}; & \node[font=\scriptsize\sffamily, text width=2.2cm] {Autoregressive (No latent)}; \\
    };
    \end{tikzpicture}
    \caption{A concise roadmap of generative models}
    \label{fig:decision_tree_landscape_vertical}
\end{figure}
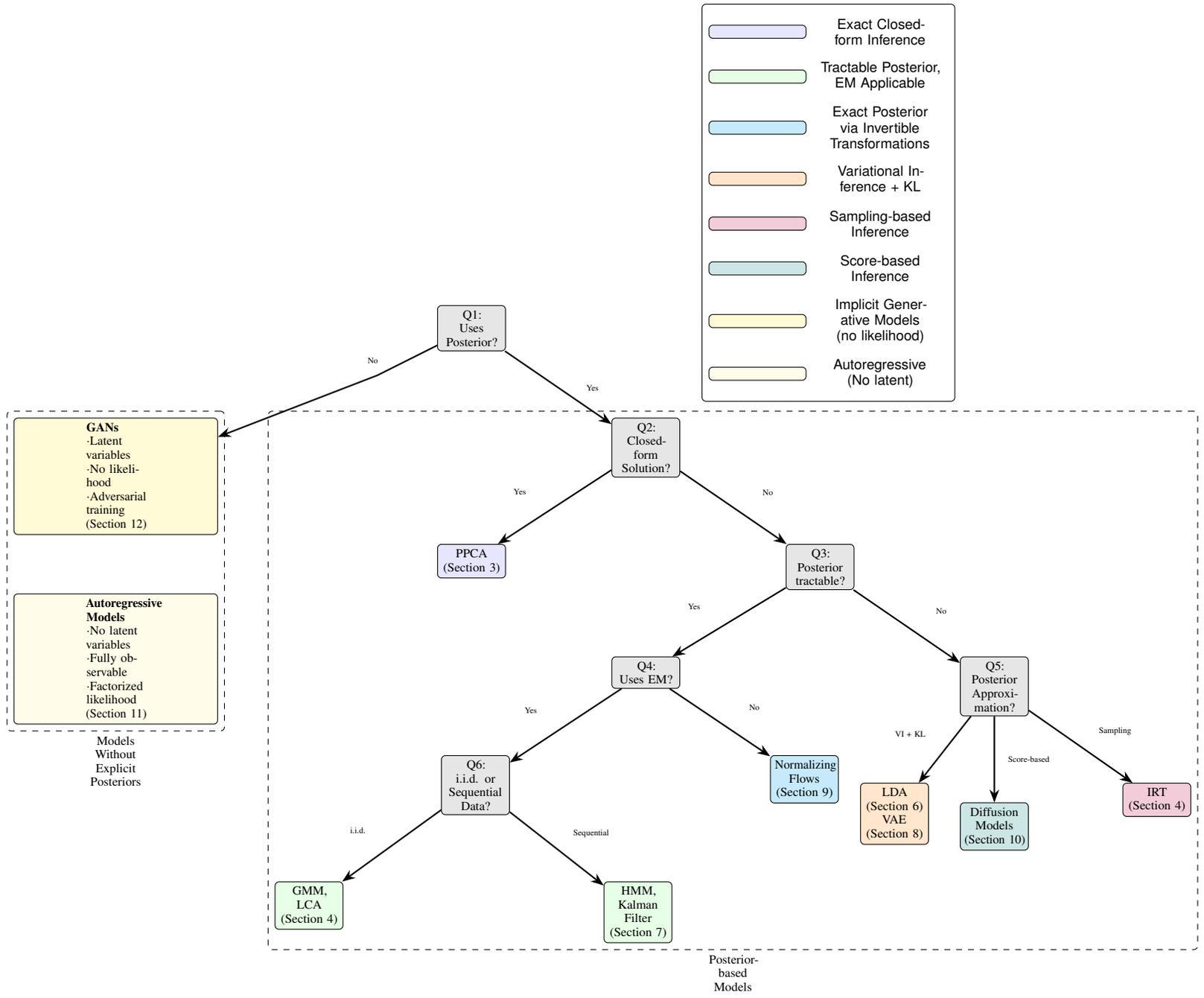
\end{landscape}

\section{Probabilistic PCA: A Continuous Latent and Continuous Observation Model}\label{sec:ppca}

This section introduces probabilistic principal component analysis (PPCA), a model characterized by continuous latent and continuous observed variables~\cite{tipping1999ppca}. As a probabilistic generalization of classical Principal Component Analysis (PCA), PPCA provides a natural and intuitive entry point into the study of latent variable models.
Here, PPCA serves as a conceptual anchor for exploring fundamental ideas in probabilistic modeling, particularly the structured relationships between observed data and latent representations, as governed by a set of core probability distributions~\cite{bishop2006, murphy2012}:
\begin{itemize}
    \item \( P(X) \) – the marginal distribution of observations.
    \item \( P(Z) \) – the prior distribution of the latent variables.
    \item \( P(X | Z) \) – the likelihood, describing how observations are generated from latent variables.
    \item \( P(Z | X) \) – the posterior distribution, representing the updated belief about the latent variables given observations.
    \item \( P(X, Z) \) – the joint distribution of both observations and latent variables.
\end{itemize}

A key focus of this discussion is the posterior distribution \( P(Z | X) \), which serves as a unifying concept across a wide range of probabilistic latent variable models~\cite{bishop1999latent}. It plays a pivotal role in inference, guiding the learning of compressed latent representations and enabling generative sampling. The principles underlying this posterior formulation form the foundation for the more advanced models introduced in subsequent sections.

\subsection{A Brief Recap of Principal Component Analysis (PCA)}

Principal Component Analysis (PCA) is a widely used method for dimensionality reduction~\cite{jolliffe2002}. Given standardized data (zero mean and unit variance), PCA identifies principal components that maximize variance retention. The variance of a dataset along a principal component \( u \) is given by:

\[
\frac{1}{N} \sum_{i} (x^{(i)^T} u)^2 = \sum_{i} u^T x^{(i)} x^{(i)^T} u.
\]

Maximizing this expression subject to \( \|u\| = 1 \) leads to the principal eigenvector of the empirical covariance matrix:

\[
\Sigma = \frac{1}{N} \sum_{i} x^{(i)} x^{(i)^T}.
\]

In practice, the lower-dimensional representation \( Z \) of the data \( X \) is often computed as:

\[
Z = W^T X,
\]

where \( X \) is arranged in columns as \([x^{(1)} x^{(2)} \dots x^{(N)}]\), with \( x^{(i)} \in \mathbb{R}^D \). The matrix \( W \) consists of the principal components (eigenvectors) \([w^{(1)} w^{(2)} \dots w^{(K)}]\), where each column represents one principal component. Since the eigenvectors of a symmetric covariance matrix are orthogonal, it follows that \( W^T = W^{-1} \), leading to: $Z = W^{-1} X.$

One possible interpretation is from the perspective of row-by-column multiplication. In this view, each row of \( W^T \) represents the coordinates of an eigenvector. When a row of \( W^T \) is dot-multiplied with a column of \( X \), it measures the similarity between the data in that dimension and the corresponding eigenvector. This operation projects the data onto the eigenvector, determining its contribution to the transformed representation and yielding the new coordinate in the eigenvector’s basis.

This result can also be interpreted as expressing \( X \) in the coordinate system defined by \( W \). In this transformation, each row of \( Z \) represents the projection of the data onto an eigenvector. Thus, the matrix \( W \) serves as a transformation matrix consisting of eigenvectors, while \( Z \) represents the lower-dimensional coordinates of the data. This interpretation facilitates the understanding for extending PCA to its probabilistic formulation in PPCA.

\subsection{From PCA to PPCA}

While PCA provides a deterministic projection, it lacks a mechanism for modeling uncertainty, handling missing data, or defining a generative process. Probabilistic PCA addresses these limitations by placing a distributional framework around the latent–observation relationship~\cite{bishop2006}.
%Probabilistic PCA extends PCA by adopting a generative probabilistic framework~\cite{bishop2006}. 
The process begins by assuming a latent variable \( Z \) sampled from a Gaussian distribution: $Z \sim \mathcal{N}(0, I), \quad Z \in \mathbb{R}^M$, where \( Z \) represents the lower-dimensional latent representation of the observed data, except that initially, it is treated as a prior assumption rather than a directly observed quantity. In reality, we only observe \( X \), so the goal is to estimate the posterior distribution \( P(Z | X) \), which incorporates additional uncertainty absent in deterministic PCA. The mean of the updated prior of $P(Z)$, i.e., the posterior, \( \mathbb{E}[Z | X] \), provides the most probable representation of the data.

To generate observations, this latent variable undergoes a linear transformation via a weight matrix \( W \in \mathbb{R}^{D \times M} \), combined with a mean vector \( \mu \), and perturbed by Gaussian noise:

\[
X = WZ + \mu + \epsilon, \quad \epsilon \sim \mathcal{N}(0, \sigma^2 I).
\]

Since for a given \( Z \), \( WZ + \mu \) is a constant, the conditional distribution \( P(X | Z) \) follows from the noise \( \epsilon \), meaning it remains Gaussian. Hence, the conditional distribution of \( X \) given \( Z \) is:
\[
P(X | Z) = \mathcal{N}(X | WZ + \mu, \sigma^2 I).
\]

The next topic of interest is to work out the expression of $P(X)$, which can be extremely useful in maximising the likelihood of seeing the observations. It would be tempting, by integrating out \( Z \), to obtain the marginal distribution:
\[
P(X) = \int P(X | Z) P(Z) dZ.
\]

At first glance, one might assume that the sum of two Gaussian distributions remains Gaussian. While this is not always true for integrals, in this case, the transformation \( X = WZ + \mu + \epsilon \) is linear. Therefore, the resulting marginal distribution \( P(X) \) remains Gaussian. Conversely, if \( X \) were a nonlinear function of \( Z \), such as \( X = f(Z) + \mu + \epsilon \) with \( f(Z) = Z^2 \), the resulting \( P(X) \) would no longer be Gaussian.

From another view, We can also easily confirm the shape of \( P(X) \) for PPCA , whereby we note that \( WZ \) is a linear transformation of a Gaussian, which remains Gaussian. Adding a shift \( \mu \) preserves Gaussianity, and adding another independent Gaussian noise \( \epsilon \) also results in a Gaussian distribution. As such we are certain that the resulting \( P(X) \) is gaussian, then the expected value and covariance of \( X \) can be computed as:

\[
\mathbb{E}[X] = \mathbb{E}[WZ + \mu + \epsilon] = W \mathbb{E}[Z] + \mathbb{E}[\mu] + \mathbb{E}[\epsilon] = \mu.
\]

\[
\text{Cov}(X) = \mathbb{E}[(X - \mathbb{E}[X])(X - \mathbb{E}[X])^T]
\]

\[
= \mathbb{E}[(WZ + \epsilon)(WZ + \epsilon)^T] = W \mathbb{E}[ZZ^T] W^T + \mathbb{E}[\epsilon \epsilon^T] = W I W^T + \sigma^2 I = WW^T + \sigma^2 I.
\]

Thus,
\[
P(X) = \mathcal{N}(X | \mu, WW^T + \sigma^2 I).
\]

With the analytical expression for \( P(X) \), we can estimate the parameters \( W \), \( \sigma^2 \), and \( \mu \) using maximum likelihood estimation (MLE). Further, With \( P(Z) \), \( P(X) \), and \( P(X | Z) \) known, for linear-gaussian models, we can directly derive the expression for the posterior distribution \( P(Z | X)\)~\cite{bishop2006}, which is to be explored further next.

\subsection{The Role of $P(Z|X)$ in Learning Latent Representations and Reconstruction}

While the MLE primarily focuses on maximizing the marginal likelihood \( P(X) \), understanding the posterior distribution \( P(Z | X) \) is crucial for learning meaningful latent representations and utilizing the model beyond likelihood estimation~\cite{tipping1999ppca, bishop2006}. 

Given that both \( P(X) \), \( P(Z) \) and \( P(X | Z) \) follow Gaussian distributions, the posterior distribution is given by:
\[
P(Z | X) = \mathcal{N}(Z | M^{-1} W^T (X - \mu), \sigma^2 M^{-1}),
\]
where \( M = W^T W + \sigma^2 I \). 
It represents the posterior distribution of the lower-dimensional representation of observations, offering insight into how data is transformed into a latent space. Revisiting the principles of PCA, where data is projected onto a lower-dimensional subspace, the key quantity of interest in PPCA is \( P(Z | X) \). Specifically, the mean of this posterior, which provides the most probable estimate of the latent representation, is given by:

\[
M^{-1} W^T (X - \mu),
\]

where \( M^{-1} W^T \) functions similarly to \( W^T \) in PCA as a transformation matrix. However, in PPCA, the term \( M^{-1} \) accounts for the uncertainty introduced by noise \( \epsilon \), which is absent in deterministic PCA. 

Thus, PPCA extends classical PCA by incorporating a probabilistic interpretation. Additionally, unlike in PCA, where \( W^T \) is typically orthogonal, in PPCA this is not necessarily the case. Furthermore, the expression \( (X - \mu) \) inherently includes the mean-centering step, eliminating the need for explicit mean-centering as a preprocessing step~\cite{bishop2006}.

From the perspective of reconstructing the original observation via its latent representation \( Z^* \), we can recover \( X \) using the generative model:
\[
X^* = W Z^* + \mu.
\]
Where substituting the expression for \( Z^* \) using its most probable estimate of the latent representation as $Z^* = M^{-1} W^T (X - \mu)$, we obtain:
\[
X^* = W M^{-1} W^T (X - \mu) + \mu.
\]

Rearranging,
\[
(X^* - \mu) = W M^{-1} W^T (X - \mu).
\]

This formulation highlights how PPCA approximates the reconstruction of \( X \): In standard PCA, where \( W \) is orthonormal and \( \sigma^2 = 0 \), the term \( M^{-1} W^T \) simplifies to \( W^T \), leading to perfect reconstruction: \( X^* = X \). whereas in PPCA, the noise variance \( \sigma^2 \) affects reconstruction, as $M^{-1} = (W^T W + \sigma^2 I)^{-1}$ is not the identity matrix. This results in an approximation:
\[
  X^* \approx X.
  \]

The reconstructed \( X^* \) is thus a smoothed probabilistic version of \( X \)~\cite{bishop2006}, rather than an exact recovery. The presence of \( \sigma^2 \) causes \( X^* \) to be a regularized reconstruction - the smaller the value of \( \sigma^2 \), the closer \( X^* \) is to \( X \).

In summary, while MLE focuses on maximizing \( P(X) \), understanding \( P(Z | X) \) is essential for practical applications in latent representation learning and inference. The posterior \( P(Z | X) \) enables the transformation of observed data \( X \) into the latent space \( Z \), facilitating feature extraction, dimensionality reduction, and the mapping of new observations for further analysis. 
In the next discussion, we will explore the role of \( P(Z | X) \) in generative sampling.

\subsection{The Role of $P(Z | X)$ in Generative Sampling}

Beyond its role in learning compressed latent representations, the posterior distribution \( P(Z | X) \) is fundamental to data generation, positioning PPCA as a generative model capable of synthesizing new observations~\cite{bishop2006}. In latent variable models, the generative process typically follows a two-step procedure: 1) A latent vector is first sampled in the latent space; 2) This latent sample is then mapped through the generative model to produce an observed data point.

In the case of PPCA, when the goal is to generate realistic variations of the training data, it may be advantageous to sample from the posterior distribution \( P(Z \mid X) \), which incorporates information from an observed data point \( X \). A latent vector drawn from this posterior is then mapped to the data space via the generative model:
\[
X = WZ + \mu + \epsilon, \quad \epsilon \sim \mathcal{N}(0, \sigma^2 I),
\]
resulting in a new synthesized observation. Sampling from this distribution facilitates that the latent variables are consistent with the observed data, thereby preserving the underlying structure of the training distribution.

Alternatively, to generate novel and diverse samples not tied to specific training examples, one may sample directly from the prior distribution. In PPCA, the prior is defined as a standard multivariate Gaussian: $P(Z) = \mathcal{N}(0, I)$.
While the prior remains fixed across datasets, one might expect that models trained on different datasets would generate similar samples. In fact, the training data influences the generated samples through the learned generative mapping \( P(X | Z) \), ensuring that generated data remains dataset-specific.

This distinction arises because: In PPCA, the learned weight matrix \( W \) and mean \( \mu \) determine how latent variables are mapped to observed data:
  \[
  X_{\text{new}} \sim \mathcal{N}(WZ + \mu, \sigma^2 I).
  \]
Since \( W \) and \( \mu \) are estimated from the training data, different datasets yield different mappings, even though \( Z \sim \mathcal{N}(0, I) \) remains the same~\cite{bishop2006, murphy2012}. Thus, while the prior \( P(Z) \) is unchanged across datasets, the generative mapping \( P(X | Z) \) ensures that generated samples retain dataset-specific characteristics.

The choice between sampling from the posterior or the prior depends on the generative objective. Posterior sampling is appropriate when generating variations of known data points, whereas prior sampling is better suited for producing entirely new instances. If the latent space has been learned effectively, the prior \( P(Z) \) can serve as a meaningful source of generative diversity. This principle underlies modern generative models such as Variational Autoencoders and Generative Adversarial Networks, where sampling from the prior enables the synthesis of new data beyond the training distribution~\cite{berahmand2024autoencoders,goodfellow2014generative}.

The mechanisms described here — defining a prior, likelihood, and analytically tractable posterior — will serve as recurring building blocks as we extend to more complex models such as GMMs, VAEs, and diffusion processes.

\section{Introduction to Alternative Latent Models}\label{sec:finite}

Building on PPCA as a representative of the continuous–continuous category, we now complete our tour of the four canonical latent–observation structures introduced in Section~\ref{sec:overview}, covering models with discrete latent variables, discrete observations, and mixed types. Each brings new modeling assumptions, inference challenges, and practical applications.

\subsection{Discrete Latent Variables – Continuous Observations: Gaussian Mixture Models}

The Gaussian Mixture Models (GMMs) is characterized by a categorical latent variable and continuous observations. They provide a probabilistic framework for clustering continuous data. Unlike k-means, which assigns hard cluster memberships, GMMs represent uncertainty through soft assignments via posterior probabilities. GMMs are widely used in applications such as speaker recognition, image segmentation, and anomaly detection~\cite{bishop2006, murphy2012}. The generative process follows these steps:

1. Sample a latent variable \( Z_k \) from a categorical distribution:

\[
P(Z_k) = \pi_k, \quad \text{where } \sum_k \pi_k = 1.
\]

Expressing this in a factorized form:
$P(Z) = \prod_k \pi_k^{z_k}$.

2. Given \( Z_k \), the corresponding observation is sampled from a Gaussian distribution with mean \( \mu_k \) and covariance \( \Sigma_k \):

\[
P(X | Z_k) = \mathcal{N}(X | \mu_k, \Sigma_k).
\]

Expressing this in a factorized form:

\[
P(X | Z) = \prod_k \mathcal{N}(X | \mu_k, \Sigma_k)^{z_k}.
\]

From the above definitions, the marginal likelihood \( P(X) \) is given by:

\[
P(X) = \sum_k P(X | Z_k) P(Z_k) = \sum_k \pi_k \mathcal{N}(X | \mu_k, \Sigma_k).
\]

As discussed in the previous section, \( P(Z | X) \) plays a crucial role in interpreting latent models. In PPCA, the posterior \( P(Z | X) \) maps continuous observations \( X \) to a lower-dimensional latent space, serving as a form of dimensionality reduction. In contrast, in GMM, \( P(Z | X) \) maps continuous observations \( X \) to a discrete latent space, effectively performing a classification or clustering task. The transformation can be understood as:

\[
X \in \mathbb{R}^D \quad \to \quad Z \in \{z_1, z_2, \dots, z_K\}.
\]

This means that instead of reducing dimensionality, GMM assigns each data point probabilistically to one of \( K \) clusters. Using Bayes' theorem, the posterior distribution \( P(Z | X) \), which determines how observations update the prior categorical distribution of \( Z \), is given by:

\[
P(Z | X) = \frac{P(X | Z) P(Z)}{P(X)} = \frac{\pi_k \mathcal{N}(X | \mu_k, \Sigma_k)}{\sum_j \pi_j \mathcal{N}(X | \mu_j, \Sigma_j)}.
\]

This posterior allows GMM to assign cluster membership probabilities to each observation rather than hard assignments. To estimate the parameters \( \mu_k \), \( \Sigma_k \), and \( \pi_k \), we maximize the log-likelihood of the observed data:

\[
\theta_{\text{MLE}} = \arg\max_{\theta} \sum_i \log \sum_k \pi_k \mathcal{N}(X^{(i)} | \mu_k, \Sigma_k).
\]

However, direct analytical optimization is intractable due to the presence of the latent variable \( Z \), which results in a log term between two summations. To overcome this, the Expectation-Maximization (EM) algorithm is employed~\cite{dempster1977maximum}. The EM algorithm iteratively: computes the expected values of the latent variables; and then maximizes the expected log-likelihood with respect to model parameters. 

This iterative approach ensures convergence to a local optimum, making GMM a powerful tool for clustering and density estimation. The EM algorithm is not only instrumental in GMMs but also serves as a fundamental optimization framework for many probabilistic latent models where direct analytical solutions are unavailable. We will explore this further in Section~\ref{sec:elbo_em}.

\subsection{Discrete Latent Variables – Discrete Observations: Latent Class Analysis}

In contrast to GMM, which clusters continuous observations into discrete categories, Latent Class Analysis (LCA) deals with scenarios where both latent variables and observed data are discrete~\cite{mccutcheon1987latent, weller2020latent}. LCA is widely used in fields such as social sciences, psychology, and education to uncover hidden categorical groupings within observed categorical responses~~\cite{collins2009latent}.

LCA assumes that an observed categorical variable \( X \) is generated from an underlying discrete latent variable \( Z \), which represents an unobserved "class" or "group" that explains the variability in \( X \), with the generative process as:

1. Sample the latent class \( Z \) from a categorical distribution:
\[
   P(Z_k) = \pi_k, \quad \text{where } \sum_k \pi_k = 1.
\]
Here, \( \pi_k \) represents the probability of an observation belonging to latent class \( k \) with the probability mass function expressed as:
   \[
   P(Z) = \prod_k \pi_k^{z_k}.
   \]

2. Given \( Z_k \), sample the observed categorical variable \( X \):
\[
   P(X | Z_k) = \prod_j P(X_j | Z_k).
\]
   Since each observed variable \( X_j \) is categorical, its probability distribution is typically modeled using a multinomial distribution~\cite{bishop2006}, where the conditional probabilities \( P(X_j | Z_k) \) describe the likelihood of each category given the latent class. Thus, the probability of observing \( X \) given \( Z \) can be expressed as:
\[
P(X | Z) = \prod_k \prod_j P(X_j | Z_k)^{z_k}.
\]

The marginal likelihood of the observations is obtained by summing over all possible latent classes:

\[
P(X) = \sum_k P(X | Z_k) P(Z_k) = \sum_k \pi_k \prod_j P(X_j | Z_k).
\]

Similar to GMM, the posterior probability \( P(Z | X) \) plays a crucial role in LCA by determining how likely an observation belongs to each latent class. However, unlike GMM, where the observations are continuous, LCA operates entirely in the discrete space. The transformation in LCA can be understood as:
\[
X \in \{x_1, x_2, \dots, x_K\} \quad \to \quad Z \in \{z_1, z_2, \dots, z_K\}.
\]

This means LCA functions as a clustering algorithm for categorical data, where each observation is probabilistically assigned to one of several discrete latent classes. Using Bayes' theorem, the posterior probability is given by:

\[
P(Z | X) = \frac{P(X | Z) P(Z)}{P(X)} = \frac{\pi_k \prod_j P(X_j | Z_k)}{\sum_l \pi_l \prod_j P(X_j | Z_l)}.
\]

To estimate the parameters \( \pi_k \) and \( P(X_j | Z_k) \), we maximize the log-likelihood of the observed data:

\[
\theta_{\text{MLE}} = \arg\max_{\theta} \sum_i \log \sum_k \pi_k \prod_j P(X_j^{(i)} | Z_k).
\]

However, direct analytical optimization is intractable due to the presence of the latent variable \( Z \), which leads to a summation within the logarithm. This makes an analytical solution infeasible. Instead, the EM algorithm is also typically applied and will be explored in greater detail in Section~\ref{sec:elbo_em}.

\subsection{Continuous Latent Variables – Discrete Observations: Item Response Theory}

Item Response Theory (IRT) models describe scenarios where the latent variable is continuous, while the observations are discrete~\cite{hambleton2013item}. This contrasts with models such as GMM and LCA, which involve discrete latent variables. IRT is widely used in psychometrics and educational measurement to model the relationship between an individual’s latent ability and their responses to test items, which are typically binary or categorical.

The generative process in a standard binary IRT model proceeds as follows:

    %Sample a continuous latent trait \( \theta \in \mathbb{R} \) from a prior distribution, where In IRT literature, the continuous latent variable is typically denoted as \( \theta \), in contrast to Z used in previous models:
1. Sample a continuous latent trait \( \theta \in \mathbb{R} \) from a prior distribution: $P(\theta) = \mathcal{N}(\theta | 0, 1)$, where in the IRT literature, the latent variable is conventionally denoted by \( \theta \), distinguishing it from the notation \( Z \) used in previous models.

2. Given \( \theta \), generate the observed binary response \( X_j \in \{0, 1\} \) for item \( j \) using a logistic model:
    \[
    P(X_j = 1 | \theta) = \sigma(a_j \theta - b_j) = \frac{1}{1 + \exp(-(a_j \theta - b_j))},
    \]
    where \( a_j \) is the item discrimination parameter, and \( b_j \) is the item difficulty parameter.

Assuming conditional independence across items, the joint likelihood of responses \( X = (X_1, \dots, X_J) \) given the latent trait \( \theta \) is:
\[
P(X | \theta) = \prod_j \sigma(a_j \theta - b_j)^{X_j} (1 - \sigma(a_j \theta - b_j))^{1 - X_j}.
\]

The marginal likelihood of the observed responses is obtained by integrating over the latent trait:
\[
P(X) = \int P(X | \theta) P(\theta) \, d\theta.
\]

In IRT, the posterior distribution \( P(\theta | X) \) maps discrete observations to a continuous latent space:
\[
X \in \{0, 1\}^J \quad \to \quad \theta \in \mathbb{R}.
\]
Using Bayes’ theorem:
\[
P(\theta | X) = \frac{P(X | \theta) P(\theta)}{P(X)} \propto P(X | \theta) P(\theta).
\]
Unlike GMM and LCA, where marginalization involves a summation over discrete latent states, the integral in \( P(X) \) does not admit a closed-form solution. Consequently, the posterior inference often requires numerical or approximate methods~\cite{blei2017vi}.

To estimate the item parameters \( a_j \) and \( b_j \), we maximize the marginal likelihood over a set of observed response vectors \( X^{(1)}, \dots, X^{(N)} \), one per individual:
\[
P(X^{(1)}, \dots, X^{(N)}) = \prod_i \int P(X^{(i)} | \theta^{(i)}) P(\theta^{(i)}) \, d\theta^{(i)}.
\]
Taking the logarithm yields the log-likelihood:
\[
\log P(X^{(1)}, \dots, X^{(N)}) = \sum_i \log \int P(X^{(i)} | \theta^{(i)}) P(\theta^{(i)}) \, d\theta^{(i)}.
\]

As in GMM and LCA, the log-likelihood function in IRT involves a nonlinear structure due to the presence of a logarithm applied to an integral over the latent variable \( \theta \). This is analogous to the log-sum formulation encountered in discrete latent models, where the marginal likelihood requires summing over latent states. While the EM algorithm remains a common approach (see Section~\ref{sec:elbo_em}), IRT models often employ alternatives such as numerical integration approaches and Bayesian methods~\cite{baker2004irt}.

Together with PPCA from the previous section, these models span the four canonical latent–observation combinations. They demonstrate how changes in the latent or observed variable types reshape the form of
$P(X)$, the structure of 
$P(Z|X)$, and the tractability of inference. This diversity motivates the general inference frameworks discussed in Section \ref{sec:elbo_em}, which unify these cases under a common optimization view.

\section{Evidence Lower Bound and Expectation Maximisation}\label{sec:elbo_em}

As previously noted, while Probabilistic PCA (PPCA) admits a closed-form solution for maximum likelihood estimation, models such as Gaussian Mixture Models (GMM), Latent Class Analysis (LCA), and Item Response Theory (IRT) do not. In these cases, the marginal log-likelihood takes the form of a logarithm applied to a sum or integral over latent variables, a nested structure that makes direct optimization analytically intractable. This motivates the focus of this section on three central topics: (1) the derivation and role of the Evidence Lower Bound (ELBO), (2) the Expectation–Maximization (EM) algorithm as a practical optimization strategy, and (3) the application of EM to the latent models introduced so far—including GMM, LCA, and IRT. Although PPCA has a closed-form solution, we include it here as an illustrative example of EM in action.

\subsection{Jensen's Inequality and Evidence Lower Bound}

%[cite Andrew's CS229 notes]

As a general scenario for parameter estimation in latent variable models, we are interested in maximizing the marginal likelihood of the observed data:
\[
\max_{\theta} p(x; \theta) = \max_{\theta} \sum_z p(x | z; \theta) p(z) = \sum_z p(x, z; \theta).
\]
This can be interpreted as marginalizing over the latent variable \( Z \). For a dataset \( \{x^{(i)}\} \), we aim to maximize the likelihood of the observations:
\[
\max_{\theta} \prod_i p(x^{(i)}; \theta) \quad \Rightarrow \quad \max_{\theta} \sum_i \log p(x^{(i)}; \theta) = \sum_i \log \sum_z p(x^{(i)}, z; \theta).
\]

However, the presence of the summation inside the logarithm generally makes this objective intractable to solve analytically. To address this, we turn to Jensen's inequality, which provides a tool for bounding transformations of expectations. Specifically, Jensen’s inequality states:
\[
f(\mathbb{E}[X]) \leq \mathbb{E}[f(X)] \quad \text{if } f \text{ is convex},
\quad \text{and} \quad
f(\mathbb{E}[X]) \geq \mathbb{E}[f(X)] \quad \text{if } f \text{ is concave}.
\]

In our case, the function of interest is the logarithm, which is both concave and monotonically increasing. Thus, we can apply the inequality:
\[
\log \mathbb{E}[f(Z)] \geq \mathbb{E}[\log f(Z)].
\]

To apply this, we first express the quantity as an expectation by introducing an auxiliary distribution \( Q(z) \) over the latent variable \( Z \), such that \( \sum_z Q(z) = 1 \).
 This allows us to rewrite the marginal likelihood for a single instance as:
\[
\log \sum_z p(x^{(i)}, z; \theta) = \log \sum_z Q(z) \frac{p(x^{(i)}, z; \theta)}{Q(z)} = \log \mathbb{E}_{z \sim Q} \left[ \frac{p(x^{(i)}, z; \theta)}{Q(z)} \right].
\]

Applying Jensen's inequality to this expression yields the following lower bound:
\[
\log \mathbb{E}_{z \sim Q} \left[ \frac{p(x^{(i)}, z; \theta)}{Q(z)} \right] \geq \mathbb{E}_{z \sim Q} \left[ \log \frac{p(x^{(i)}, z; \theta)}{Q(z)} \right] = \sum_z Q(z) \log \frac{p(x^{(i)}, z; \theta)}{Q(z)}.
\]

This provides a lower bound on the original log-likelihood, commonly referred to as Evidence Lower Bound (ELBO):
\[
\text{ELBO}(x; Q, \theta) = \sum_z Q(z) \log \frac{p(x, z; \theta)}{Q(z)}.
\]

The advantage of working with the ELBO is that the logarithm is now inside the summation, which makes the expression more amenable to optimization techniques. Ideally, we want this bound to be as tight as possible—in the best case, we want the inequality to become an equality.

To understand when the inequality becomes an equality, observe the form of the bound:
\[
\log \mathbb{E}_{z \sim Q} \left[ \frac{p(x, z; \theta)}{Q(z)} \right] \geq \sum_z Q(z) \log \frac{p(x, z; \theta)}{Q(z)}.
\]
Equality holds when the ratio \( \frac{p(x, z; \theta)}{Q(z)} \) is constant with respect to \( z \), that is:
\[
\frac{p(x, z; \theta)}{Q(z)} = c \quad \text{for all } z,
\]
for some constant \( c \). Taking the sum over both sides:
\[
\sum_z p(x, z; \theta) = \sum_z c \cdot Q(z) = c \cdot \sum_z Q(z).
\]
Since \( Q(z) \) is a valid probability distribution, \( \sum_z Q(z) = 1 \), and we have:
\[
\sum_z p(x, z; \theta) = p(x; \theta) = c.
\]
Substituting back, we find:
\[
Q(z) = \frac{p(x, z; \theta)}{p(x; \theta)} = p(z | x; \theta),
\]
which is exactly the posterior distribution of the latent variable given the observation. We can verify this by plugging this choice of \( Q(z) \) into the ELBO, which recovers the exact log-likelihood:
\[
\log \sum_z p(x, z; \theta) = \sum_z Q(z) \log \frac{p(x, z; \theta)}{Q(z)} \quad \text{when } Q(z) = p(z | x; \theta).
\]

This result establishes the ELBO as both a theoretically grounded and practical objective for optimization in latent variable models. %For a more detailed explanation of Jensen’s inequality and its application in latent variable models, see~\cite{ng2022cs229,bishop2006}
For a deeper discussion of Jensen’s inequality and its role in deriving the ELBO, refer to~\cite{ng2022cs229,bishop2006}.

\subsection{Expectation-Maximization Algorithm via ELBO}
Directly maximizing the marginal log-likelihood,$\log p(x; \theta) = \log \sum_z p(x, z; \theta),$
is generally intractable in latent variable models due to the summation over latent variables appearing inside the logarithm. This nested structure complicates differentiation and optimization with respect to the parameters \( \theta \), especially when the latent space is large or complex.

In contrast, the Evidence Lower Bound (ELBO) rewrites the objective in a more tractable form, where the log operation is moved inside an expectation. This allows the use of standard optimization tools. The ELBO serves as a lower bound on the log-likelihood: $\text{ELBO}(x; Q, \theta) = \sum_z Q(z) \log \frac{p(x, z; \theta)}{Q(z)}$, and becomes tight when \( Q(z) = p(z | x; \theta) \), the true posterior distribution. Thus, the ELBO provides a surrogate objective that is both tractable and theoretically justified.

Although the ELBO provides a tractable lower bound on the log-likelihood, directly optimizing it with respect to both \( Q(z) \) and \( \theta \) simultaneously is still challenging. This is because the ELBO is a functional of \( Q(z) \) and a parametric function of \( \theta \), and the optimal \( Q(z) \) — which makes the bound tight — is itself dependent on \( \theta \). Optimizing both at the same time would require navigating a joint space of probability distributions and model parameters, which is difficult in practice due to their entangled dependencies.

The Expectation-Maximization (EM) algorithm addresses this issue by alternating between two simpler subproblems~\cite{bishop2006,ng2022cs229,dempster1977maximum}. At each iteration, it incrementally improves the ELBO via two coordinated steps:
\begin{itemize}%[leftmargin=*]
    \item \textbf{E-step}: Fix the current parameter estimate \( \theta^{(t)} \), and compute the posterior distribution over latent variables:
    \[
    Q^{(t+1)}(z) = p(z | x; \theta^{(t)}).
    \]
    This choice of \( Q \) tightens the ELBO and ensures it equals the true log-likelihood at the current \( \theta^{(t)} \).

    \item \textbf{M-step}: Fix \( Q^{(t+1)}(z) \), and update the parameters by maximizing the ELBO with respect to \( \theta \). This reduces to maximizing the expected complete-data log-likelihood:
    \[
    \theta^{(t+1)} = \arg\max_{\theta} \sum_z Q^{(t+1)}(z) \log \frac{p(x, z; \theta)}{Q^{(t+1)}(z)}.
    \]
\end{itemize}

Each EM iteration guarantees that the ELBO does not decrease, and since the ELBO lower-bounds the true log-likelihood, this results in a monotonic improvement of the marginal likelihood \( \log p(x; \theta) \).

In principle, the ELBO can be optimized with respect to both \( Q(z) \) and \( \theta \) simultaneously using gradient-based methods. This approach is necessary when the posterior distribution \( p(z | x; \theta) \) does not admit a closed-form expression, making the E-step of EM inapplicable. In such cases, one often resorts to defining a parameterized approximation for \( Q(z) \) and jointly optimizes it along with \( \theta \). This strategy forms the basis of a broader class of methods known as variational inference, which will be introduced in Section~\ref{sec:vi}.

However, for many classical models such as GMMs and LCA, the posterior \( p(z | x; \theta) \) can be computed exactly in closed form. In these settings, the EM algorithm is often preferred due to its simplicity, interpretability, and guaranteed monotonic improvement of the log-likelihood~\cite{bishop2006}. By performing exact inference in the E-step, EM avoids the added complexity and potential instability that can arise when using gradient-based optimization over probability distributions.

\subsection{Applications of EM in Latent Variable Models}

The EM algorithm provides a unifying framework for parameter estimation in latent variable models by iteratively optimizing the ELBO. We now examine how EM applies to the probabilistic models introduced earlier, to illustrate how it provides practical solutions to otherwise intractable optimization problems in latent variable models.

\paragraph{Probabilistic PCA (PPCA).} Although PPCA has a closed-form solution via spectral decomposition, it can also be derived using the EM algorithm~\cite{tipping1999ppca,bishop2006}. This EM-based approach is especially beneficial in scenarios involving missing data or when exact eigendecomposition is computationally impractical especially in the context of big data. 

In the \textbf{E-step}, we compute the posterior distribution over the latent variables:
\[
Q(Z) = p(Z | X; \theta^{(t)}),
\]
which for PPCA remains Gaussian and has a known analytical form. Specifically, for each data point \( x^{(i)} \), the posterior is:
\[
p(z | x^{(i)}; \theta) = \mathcal{N}(z | M^{-1} W^T (x^{(i)} - \mu), \sigma^2 M^{-1}), \quad \text{where } M = W^T W + \sigma^2 I.
\]

In the \textbf{M-step}, we maximize the expected complete-data log-likelihood:
\[
\mathbb{E}_{Q(Z)} [\log p(X, Z; \theta)],
\]
which leads to updates for the parameters \( W \), \( \mu \), and \( \sigma^2 \) that align with the maximum likelihood estimates under the PPCA model.

\paragraph{Gaussian Mixture Models (GMMs).} In GMMs, the marginal log-likelihood:
\[
\log p(x) = \log \sum_k \pi_k \mathcal{N}(x | \mu_k, \Sigma_k),
\]
is intractable to optimize directly due to the summation inside the log. EM circumvents this by introducing latent variables indicating cluster membership~\cite{bishop2006,ng2022cs229}.

In the \textbf{E-step}, we compute the responsibilities:
\[
\gamma_{ik} = p(z_k = 1 | x^{(i)}; \theta) = \frac{\pi_k \mathcal{N}(x^{(i)} | \mu_k, \Sigma_k)}{\sum_j \pi_j \mathcal{N}(x^{(i)} | \mu_j, \Sigma_j)}.
\]

In the \textbf{M-step}, we maximize the ELBO by updating the parameters using weighted averages:
\begin{align*}
\pi_k^{(t+1)} &= \frac{1}{N} \sum_{i=1}^N \gamma_{ik}, \\
\mu_k^{(t+1)} &= \frac{\sum_i \gamma_{ik} x^{(i)}}{\sum_i \gamma_{ik}}, \\
\Sigma_k^{(t+1)} &= \frac{\sum_i \gamma_{ik} (x^{(i)} - \mu_k)(x^{(i)} - \mu_k)^T}{\sum_i \gamma_{ik}}.
\end{align*}

\paragraph{Latent Class Analysis (LCA).} LCA is similar in structure to GMM but is used for discrete observed variables. Each observation is assumed to belong to one of \( K \) latent classes, and the observed categorical responses are conditionally independent given the latent class~\cite{mccutcheon1987latent}.

In the \textbf{E-step}, we compute the posterior class probabilities for each observation:
\[
\gamma_{ik} = p(z_k = 1 | x^{(i)}; \theta) = \frac{\pi_k \prod_j P(X_j^{(i)} | z_k)}{\sum_l \pi_l \prod_j P(X_j^{(i)} | z_l)}.
\]

In the \textbf{M-step}, we update the model parameters as follows:
\begin{align*}
\pi_k^{(t+1)} &= \frac{1}{N} \sum_i \gamma_{ik}, \\
P(X_j = c | z_k)^{(t+1)} &= \frac{\sum_i \gamma_{ik} \cdot \mathbb{I}(X_j^{(i)} = c)}{\sum_i \gamma_{ik}},
\end{align*}
where \( \mathbb{I}(\cdot) \) is the indicator function.

\medskip

\paragraph{Item Response Theory (IRT).} In IRT, the goal is to estimate item parameters (e.g., discrimination and difficulty) from discrete responses, under the assumption of an underlying continuous latent trait. The marginal likelihood involves an integral over the latent variable:
\[
\log p(x^{(i)}) = \log \int p(x^{(i)} | \theta) p(\theta) \, d\theta,
\]
which is analytically intractable, making direct maximization of the log-likelihood difficult~\cite{baker2004irt}.

In the \textbf{E-step}, we compute the posterior distribution over the latent trait \( \theta \) for each individual:
\[
Q(\theta^{(i)}) = p(\theta^{(i)} | x^{(i)}; \theta^{(t)}),
\]
which typically requires approximation via numerical integration, such as Gauss-Hermite quadrature, since there is no closed-form expression.

In the \textbf{M-step}, we maximize the expected complete-data log-likelihood with respect to the item parameters:
\[
\mathbb{E}_{Q(\theta)} [\log p(X, \theta; \theta_{\text{item}})],
\]
which leads to updates for discrimination and difficulty parameters based on expected sufficient statistics under the current posterior approximation.

Unlike GMM and LCA, where the E-step involves computing probabilities over a finite set of latent states, IRT requires integrating over a continuous latent space. Nevertheless, EM remains a powerful estimation technique, particularly when combined with efficient numerical integration. %It is widely used in marginal maximum likelihood estimation of IRT models, and serves as a bridge to Bayesian estimation methods when MCMC is employed in the E-step.

\paragraph{Summary.} Across all four models, EM provides an efficient and tractable framework for parameter estimation by alternating between inferring latent structure and updating model parameters. In PPCA, both steps are analytically tractable; in GMM and LCA, they involve closed-form updates over discrete latent variables; and in IRT, the E-step requires approximating integrals over continuous latent traits. Despite these differences, the core principle is unchanged: EM optimizes the ELBO and guarantees a non-decreasing likelihood at each iteration. This makes it a versatile and widely used method for probabilistic modeling with latent variables, spanning both discrete and continuous settings. When the posterior 
$p(z∣x;θ)$ is intractable, EM cannot be implemented in closed form, prompting the use of approximate inference methods such as Variational Inference (Section~\ref{sec:vi}), which extend the ELBO framework to parameterized posterior families.

\section{KL Divergence, Variational Inference, and Latent Dirichlet Allocation}\label{sec:vi}

The use of the ELBO as a surrogate objective depends critically on whether the posterior distribution \( P(z \mid x) \) can be computed in closed form. In models such as PPCA, GMM, and LCA, this is indeed the case, allowing the ELBO to be maximized exactly via the EM algorithm.

A notable intermediate case is IRT. Here, the posterior \( p(\theta^{(i)} \mid x^{(i)}; \theta) \) involves an integral over a continuous latent variable and has no analytical form. Nevertheless, EM remains applicable: the E-step can be approximated using numerical integration, while the M-step updates stay exact. This makes IRT an example of a \textit{hybrid approach}, where posterior inference is approximate but still embedded within the EM framework---avoiding the need for fully variational or sampling-based methods.

In more complex models, however, the posterior \( P(z \mid x) \) can be completely intractable, whether due to non-conjugacy, high dimensionality, or intricate dependencies. In these cases, two broad families of approximate inference strategies are typically used:

\begin{enumerate}%[leftmargin=*]
    \item \textbf{Sampling-based methods}, e.g., Monte Carlo or Gibbs sampling, draw samples from the true posterior to approximate expectations. They are asymptotically exact but often slow to converge, especially in large-scale settings.
    \item \textbf{Variational inference}, instead of sampling, which are asymptotically exact but often slow, is approximate but scalable, turning inference into an optimization problem by minimizing the divergence from the true posterior.
    %one posits a tractable family \( Q(z) \) and optimizes its parameters to minimize the divergence from the true posterior. This converts inference into an optimization problem, trading some accuracy for scalability and efficiency, particularly in large or deep probabilistic models.
\end{enumerate}

Variational inference has therefore become a central tool in both classical and modern probabilistic modeling. In the remainder of this section, we first introduce the Kullback--Leibler divergence as a measure of dissimilarity between distributions, then formalize the variational inference framework through the example of Latent Dirichlet Allocation, a widely used model for uncovering latent structure in discrete data such as text.

\subsection{Entropy, Cross Entropy and KL Divergence}

To perform variational inference, we must define a way to measure how close a proposed distribution \( Q(z) \) is to the true posterior \( P(z \mid x) \). This is typically done using the Kullback--Leibler (KL) divergence~\cite{goodfellow2016deep}. To understand this, we first revisit the concept of entropy and related quantities:

\textbf{Entropy} \( H(P) \) measures the uncertainty inherent in a probability distribution \( P \). For a discrete random variable \( z \sim P(z) \), the entropy is defined as:
    \[
    H(P) = -\sum_z P(z) \log P(z).
    \]
    This represents the expected amount of information (or surprise) from drawing samples according to \( P \). Higher entropy indicates more unpredictability.

\textbf{Cross-entropy} \( H(P, Q) \) measures the expected number of bits required to encode samples from \( P \) using a coding scheme optimized for \( Q \), and is defined as:
    \[
    H(P, Q) = -\sum_z P(z) \log Q(z).
    \]
    If \( Q \) is close to \( P \), the cross-entropy will be close to the entropy of \( P \). However, if \( Q \) diverges from \( P \), more bits are needed on average.

\textbf{Kullback--Leibler (KL) divergence} quantifies the difference between two probability distributions \( Q \) and \( P \), and is defined as:
    \[
    D_{\text{KL}}(Q \| P) = \sum_z Q(z) \log \frac{Q(z)}{P(z)}.
    \]
    This can also be written as the difference between cross-entropy and entropy:
    \[
    D_{\text{KL}}(Q \| P) = H(Q, P) - H(Q).
    \]
    KL divergence measures the inefficiency (extra surprise) of assuming the distribution is \( P \) when the true distribution is \( Q \). It is always non-negative, and equal to zero if and only if \( Q = P \). In the context of variational inference, the KL divergence is used to minimize the discrepancy between the approximate posterior \( Q(z) \) and the true posterior \( P(z \mid x) \), enabling tractable inference in otherwise intractable Bayesian models.

\subsection{Latent Dirichlet Allocation}

With this background, we now demonstrate variational inference using another important latent probabilistic model—Latent Dirichlet Allocation (LDA), a generative probabilistic model designed to uncover the hidden thematic structure within a corpus of documents~\cite{blei2003latent}. Its core assumption is that each document is a mixture of latent topics, and each topic corresponds to a distribution over the vocabulary. LDA is widely used in natural language processing for topic modeling and in information retrieval for uncovering latent semantic structures in large text corpora, with key advantages lying in its ability to model documents as mixtures of topics that scale well to large datasets~\cite{chauhan2021topic}.

The generative process for LDA proceeds as follows:

\begin{enumerate}
    \item For each topic \( k \in \{1, \dots, K\} \), sample a word distribution over the vocabulary:
    \[
    \phi_k \sim \text{Dirichlet}(\beta), \quad \phi_k \in \Delta^{V-1},
    \]
    where \( \phi_k \) defines the categorical distribution over the vocabulary for topic \( k \).
    
    \item For each document \( d \in \{1, \dots, D\} \):
    \begin{enumerate}
        \item Sample the topic proportions:
        \[
        \theta_d \sim \text{Dirichlet}(\alpha), \quad \theta_d \in \Delta^{K-1}.
        \]
        \item For each word position \( n \in \{1, \dots, N_d\} \):
        \begin{itemize}
            \item Sample a topic assignment:
            \[
            z_{d,n} \sim \text{Categorical}(\theta_d).
            \]
            \item Sample a word from the selected topic's distribution:
            \[
            w_{d,n} \sim \text{Categorical}(\phi_{z_{d,n}}).
            \]
        \end{itemize}
    \end{enumerate}
\end{enumerate}

This formulation bears some conceptual similarity to Latent Class Analysis (LCA) as previously discussed. Both LCA and LDA involve latent discrete variables that influence observed categorical outcomes. However, the crucial structural difference is that LCA assigns each observation (e.g., a document) to a single latent class, from which all observations are generated. This leads to the generative path \( z \rightarrow x \), with all words drawn from a single topic-specific distribution.

In contrast, LDA follows a hierarchical process where each document is first associated with a distribution over topics, and each word is assigned its own topic from this distribution. The corresponding path is \( \phi, \theta \rightarrow z \rightarrow w \). This deeper structure allows each document to blend multiple topics, reflecting real-world textual data more naturally. The word distributions \( \phi_k \) and topic proportions \( \theta_d \) are latent variables drawn from Dirichlet priors, making exact posterior inference intractable.

\subsection{Joint Posterior and Variational Inference in LDA}

Unlike flat latent variable models such as GMM or LCA, where inference primarily focuses on estimating the posterior of a single latent variable \( z \) given the data, i.e., $p(z \mid x)$, LDA involves a more complex hierarchical structure with three layers of latent variables:
 \( \theta_d \), document-specific topic proportions; \( z_{d,n} \), word-level topic assignments; \( \phi_k \), topic-specific word distributions. All of these variables are unobserved and coupled through the generative process. The primary goal in LDA is to uncover the hidden topic structure of a corpus by estimating these latent variables based on the observed words \( w \). Formally, the inference task is to compute the joint posterior distribution:
\[
p(\theta, \phi, z \mid w, \alpha, \beta),
\]

However, computing this posterior exactly is infeasible. The marginal likelihood of the data:
\[
p(w \mid \alpha, \beta) = \int \sum p(w, \theta, \phi, z \mid \alpha, \beta) \, d\theta \, d\phi,
\]
requires summing over all possible topic assignments and integrating over all topic proportions and topic-word distributions. For a corpus with \( D \) documents, each containing \( N_d \) words and \( K \) topics, the number of possible topic assignments grows exponentially:
\[
\text{Number of combinations} = \prod_{d=1}^D K^{N_d}.
\]
Additionally, the dependency between \( \theta \), \( \phi \), and \( z \) makes it impossible to factorize the integral easily.

To overcome this challenge, variational inference introduces a family of simpler, tractable distributions \( q(\theta, \phi, z) \) to approximate the true posterior:
\[
q(\theta, \phi, z) \approx p(\theta, \phi, z \mid w).
\]

The standard approach assumes a mean-field factorization:
\[
q(\theta, \phi, z) = q(\phi) \prod_d q(\theta_d) \prod_{n=1}^{N_d} q(z_{d,n}),
\]
which breaks the complex dependencies between variables and makes optimization feasible.

The variational parameters are optimized by minimizing the KL divergence:
\[
\text{KL}(q(\theta, \phi, z) \, \| \, p(\theta, \phi, z \mid w)),
\]
which is equivalent to maximizing the ELBO:
\begin{align*}
\log p(w) &= \log \int \sum q(\theta, \phi, z) \frac{p(\theta, \phi, z, w)}{q(\theta, \phi, z)} \, d\theta \, d\phi \\
&\geq \mathbb{E}_{q} \left[ \log p(w, \theta, \phi, z) \right] - \mathbb{E}_{q} \left[ \log q(\theta, \phi, z) \right] = \text{ELBO}.
\end{align*}

Since \( \log p(w) = \text{ELBO} + \text{KL}(q \| p) \), maximizing the ELBO brings the variational distribution \( q \) closer to the true posterior. This turns inference into an optimization problem that can be solved using more scalable methods like stochastic variational inference. Variational inference thus offers a principled and efficient framework for approximate Bayesian inference in LDA, especially suited to large-scale document collections where exact methods like Gibbs sampling may become impractical.

Finally, it is worth emphasizing that LDA is formulated from a fully Bayesian perspective, in contrast to earlier models such as PPCA, GMM, and LCA, which are typically trained via maximum likelihood estimation. LDA places Dirichlet priors on both the per-document topic distributions \( \theta_d \) and the per-topic word distributions \( \phi_k \), treating them as random variables rather than fixed parameters. This Bayesian formulation enables principled uncertainty quantification, which is particularly beneficial in low-data or sparse settings---an advantage not shared by purely MLE-based models. However, in practice, the common inference strategies for LDA (such as variational inference or collapsed Gibbs sampling) still yield solutions that can be interpreted as maximizing an ELBO, making them closely related in spirit to maximum likelihood estimation despite the model's Bayesian framing.

\section{Extensions to Sequential Data: Time-Series Probabilistic Latent Variable Models}
\label{sec:time_series}

Up to this point, our discussion has focused on flat probabilistic latent variable models (PLVMs), where observations are assumed to be independent and identically distributed (i.i.d.). Examples such as probabilistic PCA, Gaussian mixture models, and latent Dirichlet allocation fit neatly into this category. However, many real-world datasets are inherently sequential, exhibiting temporal dependencies that flat models cannot capture. Time-series PLVMs extend the latent-variable framework to account for such dependencies, making them indispensable in domains such as speech recognition, finance, physiological signal analysis, motion tracking, and robotics. 

Before turning to modern deep PLVMs, we briefly review three classical time-series latent models, showing how the same probabilistic principles—latent variables, structured conditional dependencies, and EM-based learning—naturally extend to the temporal setting.

\paragraph{Hidden Markov Models (HMMs).}  
HMMs~\cite{bouguila2022hidden} extend probabilistic latent variable modeling to sequential data by introducing a sequence of latent states \( z_{1:T} = \{z_1, \ldots, z_T\} \) that evolve over time according to a first-order Markov process. Each observed variable \( x_t \) is generated conditionally on the current hidden state \( z_t \), forming a temporal model suitable for structured sequences. HMMs have been widely used in applications such as speech recognition, biological sequence analysis, and part-of-speech tagging~\cite{mor2021systematic}.

The generative process begins with sampling an initial latent state from \( p(z_1) = \pi_{z_1} \), followed by state transitions via a transition matrix \( A \in \mathbb{R}^{K \times K} \), where \( p(z_t \mid z_{t-1}) = A_{z_{t-1}, z_t} \) for \( t \geq 2 \). Given the latent state at each time step, the corresponding observation is emitted from a categorical distribution, \( p(x_t \mid z_t) = B_{z_t}(x_t) \), where \( B_{z_t} \) defines the emission probabilities over the observation space.

The joint distribution is:
\[
p(x_{1:T}, z_{1:T}) = p(z_1) \prod_{t=2}^T p(z_t \mid z_{t-1}) \prod_{t=1}^T p(x_t \mid z_t).
\]

The key inferential task is computing the posterior \( p(z_{1:T} \mid x_{1:T}) \), which provides the latent state sequence most consistent with the observed data. Inference is made tractable via the Forward-Backward algorithm, a dynamic programming technique~\cite{puterman2014markov} that efficiently computes the required posterior marginals.

Parameter estimation is typically performed using the EM algorithm. While the structure mirrors that of flat models such as GMMs and LCA, the E-step in HMMs involves computing expectations over time-dependent latent states using Forward-Backward, rather than simple marginal posteriors. The M-step updates the initial state distribution \( \pi \), transition matrix \( A \), and emission probabilities \( B \) to maximize the expected complete-data log-likelihood. %This temporal adaptation of EM makes HMMs a foundational model for sequential latent structure, albeit with limitations in modeling long-range dependencies or high-dimensional data.

\paragraph{Gaussian Hidden Markov Models (Gaussian HMMs).}
The Gaussian HMM~\cite{bouguila2022hidden} generalizes the classical discrete HMM by allowing the observed variables to be continuous-valued. While the latent variables \( z_{1:T} = \{z_1, \ldots, z_T\} \) remain discrete and evolve according to a first-order Markov chain, each observation \( x_t \in \mathbb{R}^d \) is drawn from a multivariate Gaussian distribution conditioned on the current latent state.

The generative process proceeds as follows: the initial latent state is sampled from a categorical distribution \( p(z_1) = \pi_{z_1} \), and the subsequent states follow a Markov transition matrix \( A \in \mathbb{R}^{K \times K} \), such that \( p(z_t \mid z_{t-1}) = A_{z_{t-1}, z_t} \) for \( t \geq 2 \). Each observed variable \( x_t \) is then sampled from a Gaussian distribution specific to the current state: $p(x_t \mid z_t) = \mathcal{N}(x_t; \mu_{z_t}, \Sigma_{z_t})$, where \( \mu_{z_t} \) and \( \Sigma_{z_t} \) are the state-dependent mean vector and covariance matrix, respectively.

Inference in Gaussian HMMs entails computing the posterior distribution over the latent sequence, \( p(z_{1:T} \mid x_{1:T}) \), based on the observed continuous data. This is efficiently achieved using the Forward-Backward algorithm, which remains applicable with Gaussian emission probabilities. The core structure of the algorithm remains unchanged from the discrete case, but the emission likelihoods are computed using Gaussian densities rather than categorical probabilities.

Parameter estimation also follows the EM framework. In the E-step, the Forward-Backward algorithm is used to compute the expected sufficient statistics involving marginal and pairwise posteriors. In the M-step, these are used to update the model parameters: the initial distribution \( \pi \), the transition matrix \( A \), and the emission parameters \( \mu_k \) and \( \Sigma_k \) for each latent state \( k \). %This extension makes Gaussian HMMs particularly useful in applications involving real-valued sequential data, such as speech, motion trajectories, and sensor streams~\cite{mor2021systematic}

\paragraph{Linear Dynamical Systems (LDS)} The LDS, also known as Kalman Filters~\cite{khodarahmi2023review} in the state estimation literature, extend probabilistic latent variable modeling to the continuous and linear-Gaussian setting. In contrast to HMMs and Gaussian HMMs, which assume discrete latent variables, LDS models latent dynamics and observations as continuous multivariate Gaussian processes, making them particularly well-suited for real-valued time-series data such as in control systems, robotics, and signal processing~\cite{auger2013industrial}.

In LDS, the latent state \( z_t \in \mathbb{R}^{d_z} \) evolves over time via a linear dynamical system with Gaussian noise: $z_t = A z_{t-1} + u_t, \quad u_t \sim \mathcal{N}(0, Q)$, where \( A \) is the state transition matrix and \( Q \) is the process noise covariance. The initial state \( z_1 \sim \mathcal{N}(\mu_0, \Sigma_0) \). Observations \( x_t \in \mathbb{R}^{d_x} \) are generated as linear projections of the latent states with added Gaussian noise: $x_t = C z_t + v_t, \quad v_t \sim \mathcal{N}(0, R)$, where \( C \) is the observation matrix and \( R \) is the observation noise covariance.

Given the linear-Gaussian structure, the posterior \( p(z_{1:T} \mid x_{1:T}) \) is itself Gaussian and can be computed exactly. Inference proceeds in two stages:
the Kalman Filter performs a forward recursion to compute filtered marginals \( p(z_t \mid x_{1:t}) \) sequentially; the Kalman Smoother then executes a backward pass to obtain smoothed posteriors \( p(z_t \mid x_{1:T}) \), incorporating both past and future observations.

Parameter estimation in LDS is also typically performed using the EM algorithm, as in many other latent variable models. In the E-step, the Kalman Smoother computes the expected sufficient statistics of the latent states. The M-step then updates the model parameters \( A, C, Q, R \), and the initial state parameters \( \mu_0, \Sigma_0 \) by maximizing the expected complete-data log-likelihood.

\paragraph{Summary.}  
Time-series PLVMs extend the core latent-variable framework to sequential data, replacing the i.i.d.\ assumption with latent Markov processes—discrete in HMMs and continuous in LDS. Despite introducing temporal dependencies, both classes admit exact inference through dynamic programming methods such as Forward–Backward or Kalman filtering/smoothing, and can be trained efficiently via the EM algorithm.

The structural elements of HMMs and LDS—latent state transitions, emission models, and recursive inference—form the conceptual backbone of many modern deep sequential models. Architectures such as recurrent neural network (RNN) VAEs, neural state-space models, and certain diffusion-based approaches can be seen as high-capacity, non-linear generalizations of these classical frameworks. In these modern variants, neural networks replace simple discrete or linear mappings in the transition and emission functions, while approximate inference methods (e.g., variational or amortized inference) supplant exact dynamic programming.

This progression from classical to deep sequential PLVMs reflects a broader trend that underpins the remainder of this paper: the adaptation of foundational probabilistic principles to high-dimensional, complex data through neural parameterization. The following sections introduce key families of deep generative models—beginning with Variational Autoencoders—that inherit, extend, or transform these probabilistic ideas into the state-of-the-art architectures driving modern generative AI.

\section{Variational Autoencoders: Deep Probabilistic Latent Variable Models}\label{sec:vae}

Having established how classical PLVMs extend to sequential data, we now turn to the first major class of modern deep generative models: Variational Autoencoders (VAEs). VAEs preserve the probabilistic latent-variable structure but replace the restricted, often linear or discrete parameterizations of classical models with neural networks capable of modeling complex, high-dimensional, and non-linear relationships between latent and observed variables.

The models discussed so far, i.e., PPCA, GMM, LCA, IRT, LDA, and their time-series extensions—demonstrate the breadth of the PLVM framework in its classical form, where tractable inference is ensured through simplifying assumptions. While these models are powerful and interpretable, their capacity to represent highly intricate data distributions is inherently limited. Modern deep generative models lift these constraints by parameterizing both the generative and inference components with flexible neural architectures, while retaining the core probabilistic elements such as the priors, likelihoods, and ELBO-based inference.

Within this landscape, the VAE stands as one of the most influential examples of a deep probabilistic latent variable model. It generalizes the variational inference principles seen in models like LDA to continuous, non-linear settings, providing a natural bridge to the subsequent architectures we explore—Normalizing Flows, Diffusion Models, Autoregressive Models, and GANs.

\subsection{From LDA to Variational Autoencoder}

In the previous section on Latent Dirichlet Allocation (LDA), we saw how variational inference and KL divergence provide a scalable framework for approximating intractable posteriors. These same ideas extend naturally to modern deep generative models, most notably the Variational Autoencoder (VAE)~\cite{kingma2022autoencodingvariationalbayes}, which has become a cornerstone for representation learning, dimensionality reduction, and data generation~\cite{liang2024survey,berahmand2024autoencoders}.

Like LDA, the VAE is a latent variable model with both observed and hidden variables, but it departs in two crucial ways: 1) The latent variables are continuous, typically Gaussian-distributed, rather than discrete; 2) The likelihood and variational posterior are parameterized by deep neural networks, allowing highly flexible, non-linear mappings between latent and observed spaces.

Formally, the VAE consists of: An encoder (inference network) that maps input \( x \) to a distribution over latent variables \( z \); A decoder (generative network) that maps latent variables \( z \) back to a distribution over data \( x \).

Traditional autoencoders also map data to a lower-dimensional representation and back, but do so deterministically and without a generative probabilistic interpretation. VAEs, by contrast, define a full generative process:
\[
z \sim p(z), \quad x \sim p_\theta(x \mid z),
\]
enabling the generation of new samples by first drawing from the prior \( p(z) \) and then decoding.

Where LDA’s hierarchy is defined by document-topic proportions, word-topic assignments, and observed words, the VAE’s hierarchy arises from multiple layers of non-linear transformations in the encoder and decoder. This flexibility allows VAEs to model extremely complex data distributions, but also renders the posterior \( p(z \mid x) \) analytically intractable, as discussed next.

\subsection{Intractable Posterior in VAEs}

In both LDA and VAE, the need for variational inference arises from the intractability of computing posteriors directly $p(z \mid x) = \frac{p(x \mid z)p(z)}{p(x)}$. While LDA’s complexity stems from the coupling of discrete latent variables in a probabilistic graphical model, the VAE's complexity arises from the use of deep neural networks to parameterize the encoder and decoder, which introduce multiple layers of non-linear transformations. %In both cases, the KL divergence and variational inference serve as key tools to make learning and inference feasible. 
To further understand why the posterior becomes analytically intractable in VAEs, let us explicitly describe the full generative and inference processes, including the role of neural networks and the associated probabilistic quantities.

\paragraph{1) The prior distribution of the latent variable.}
In VAEs, the latent variable \( z \in \mathbb{R}^d \) is assumed to follow a simple prior, typically a standard multivariate Gaussian: $p(z) = \mathcal{N}(z; 0, I)$, where \( I \) is the identity matrix. This acts as a regularization prior over the latent space and defines the source distribution from which latent codes are to be drawn.

\paragraph{2) The encoder: computing an approximate posterior.}
The encoder network maps the observed data \( x \in \mathbb{R}^n \) to the parameters of an approximate posterior distribution:
\[
q_{\phi}(z \mid x) = \mathcal{N}(z; \mu_{\phi}(x), \Sigma_{\phi}(x)),
\]
where \( \mu_{\phi}(x) \) and \( \Sigma_{\phi}(x) \) are functions of \( x \) computed via multiple layers of neural networks parameterized by \( \phi \); and \( \Sigma_{\phi}(x) \) is often assumed to be diagonal as \( \Sigma_{\phi}(x) = \text{diag}(\sigma^2_{\phi}(x)) \).
In other words, the encoder maps the input \( x \) through a series of non-linear transformations to produce the parameters \( \mu_{\phi}(x) \) and \( \sigma_{\phi}^2(x) \) of the distribution over latent variables.

\paragraph{3) Sampling latent variable and reparameterization trick.} A latent code \( z \) is sampled from approximate posterior:
\[
z \sim q_{\phi}(z \mid x).
\]

To make the sampling operation differentiable (since sampling from a probability distribution is a non-differentiable operation), the so-called reparameterization trick is employed:
\[
z = \mu_{\phi}(x) + \sigma_{\phi}(x) \odot \epsilon, \quad \epsilon \sim \mathcal{N}(0, I),
\]
where \( \odot \) denotes element-wise multiplication. This allows gradients to flow through \( \mu_{\phi} \) and \( \sigma_{\phi} \) during optimization.

\paragraph{4) The decoder: generating data from \( z \).}
The sampled latent variable \( z \) is passed through the decoder network, which defines the conditional distribution:
\[
p_{\theta}(x \mid z) = \text{Likelihood Model}(f_{\theta}(z)),
\]
where \( f_{\theta}(z) \) is a deep neural network parameterized by \( \theta \) and outputs the parameters of the distribution over \( x \) (e.g., mean and variance if \( x \) is continuous, or probabilities if \( x \) is categorical). The decoder may consist of multiple layers of non-linear transformations:
\[
z \rightarrow \text{Nonlinear Layer}_1 \rightarrow \cdots \rightarrow \text{Nonlinear Layer}_L \rightarrow \text{Output distribution over } x.
\]

From the above, we can tell that
the true posterior:
\[
p(z \mid x) = \frac{p(x \mid z) p(z)}{p(x)}
\]
is intractable because: 1) The likelihood \( p(x \mid z) \) is modeled by a neural network, making it non-linear and often without closed-form integration over \( z \); 2) The marginal likelihood: $p(x) = \int p(x \mid z)p(z) dz$ is analytically intractable since the integration involves the non-linear decoder network \( f_{\theta}(z) \).
%This situation is fundamentally analogous to LDA, where the coupling between \( \theta \), \( z \), and \( \phi \) makes the posterior intractable. In VAEs, the intractability is due to the \textbf{composition of multiple non-linear functions} performed by deep neural networks in both the encoder and decoder.
Thus, VAE resorts to variational inference by introducing the variational distribution \( q_{\phi}(z \mid x) \) and minimizing the KL divergence between \( q_{\phi}(z \mid x) \) and \( p(z \mid x) \) to make learning tractable, as will be discussed next.

\subsection{Variational Inference and KL Divergence for Optimizing VAEs}

Having introduced the encoder, decoder, and the reparameterization trick, we now turn to the key inference and learning challenge: how to optimize the parameters of the encoder (\( \phi \)) and decoder (\( \theta \)) given only observed data \( x \), while bypassing the intractability of the true posterior \( p(z \mid x) \).

Since the true posterior \( p(z \mid x) \) is intractable due to the non-linearity of the decoder, VAEs introduce a variational approximation:
\[
q_{\phi}(z \mid x) \approx p(z \mid x),
\]
where \( q_{\phi}(z \mid x) = \mathcal{N}(z; \mu_{\phi}(x), \Sigma_{\phi}(x)) \) is parameterized by the encoder network. This distribution is designed to be computationally tractable, enabling efficient sampling and evaluation.

Instead of maximizing the true marginal likelihood \( p(x) \) directly, VAEs maximize the Evidence Lower Bound (ELBO):
\[
\log p(x) \geq \mathbb{E}_{q_{\phi}(z \mid x)} [\log p_{\theta}(x \mid z)] - D_{\mathrm{KL}}(q_{\phi}(z \mid x) \| p(z)).
\]
Where the first term $ \mathbb{E}_{q_{\phi}(z \mid x)}[\log p_{\theta}(x \mid z)]$ is the reconstruction term, encouraging the decoder to accurately reconstruct the data \( x \) given the latent variable \( z \) sampled from the approximate posterior. The second term $ D_{\mathrm{KL}}(q_{\phi}(z \mid x) \| p(z))$ is the regularization term,  penalising divergence between the approximate posterior \( q_{\phi}(z \mid x) \) and the prior \( p(z) \). The idea is to prevent the encoder from deviating too far from the prior distribution, thus encouraging the latent space to be well-structured and enables smooth sampling.

Although VAEs are framed as Bayesian models—placing a prior over \( z \) and a likelihood over \( x \)—in practice their parameters are learned by maximizing the ELBO. This procedure is conceptually akin to maximum likelihood estimation augmented with a KL-based regularization term. Formally, the VAE training objective is:
\[
\max_{\theta, \phi} \; \mathcal{L}(\theta, \phi; x) 
= \mathbb{E}_{q_{\phi}(z \mid x)}[\log p_{\theta}(x \mid z)] 
- D_{\mathrm{KL}}(q_{\phi}(z \mid x) \| p(z)),
\]
where the first term encourages faithful reconstruction of \( x \) from \( z \), and the second penalizes divergence between the approximate posterior and the prior. With both encoder (\( \phi \)) and decoder (\( \theta \)) parameterized by deep neural networks, this objective is optimized via stochastic gradient-based methods and their variants.

Overall, the variational inference provides the theoretical foundation for optimizing VAEs. The KL divergence quantifies the approximation error of the variational posterior, and the ELBO balances reconstruction quality and latent space regularization. This framework enables VAEs to perform both inference (estimating latent variables for given data) and generation (producing new samples) within a unified probabilistic model.

\section{Normalizing Flows: Tractable Probabilistic Latent Variable Models}\label{sec:nf}

Normalizing Flows (NFs)~\cite{kobyzev2020normalizing,papamakarios2021normalizing} occupy a unique position within the PLVM family: they retain a latent-variable structure but achieve fully tractable inference and likelihood computation. Whereas deep PLVMs such as VAEs rely on variational approximations to handle intractable posteriors, NFs sidestep the issue entirely by constraining the generative mapping from latent space to data space to be invertible and differentiable. This bijective structure ensures that the posterior can be computed exactly via the inverse transformation, and that the likelihood can be evaluated precisely using the change-of-variable formula.

From the PLVM perspective, NFs define a joint distribution where the posterior \( p(\mathbf{z}_0 \mid \mathbf{x}) \) is deterministic and tractable, given by the inverse transformation \( f_\theta^{-1}(\mathbf{x}) \). The generative and inference processes are thus two sides of the same transformation, eliminating the need for variational approximations or EM-based learning. These properties make NFs particularly attractive for tasks such as density estimation, anomaly detection, and high-dimensional generative modeling. They have been successfully applied across domains including image and audio synthesis, molecular modeling, and simulation, where exact likelihood evaluation, stable training, and efficient sampling are highly desirable.

\subsection{Change of Variables and Exact Likelihood in Normalizing Flows}

Normalizing Flows are built upon a fundamental result from probability theory: the change-of-variable formula, which describes how probability densities transform under invertible and differentiable mappings. Let \( \mathbf{z} \in \mathbb{R}^d \) be a latent variable drawn from a known distribution \( p_Z(\mathbf{z}) \), and let \( \mathbf{x} = f_\theta(\mathbf{z}) \) be a smooth, bijective transformation. Then the resulting density over \( \mathbf{x} \) is given by:
\[
p_X(\mathbf{x}) = p_Z(f_\theta^{-1}(\mathbf{x})) \left| \det \left( \frac{\partial f_\theta^{-1}}{\partial \mathbf{x}} \right) \right|.
\]
Equivalently, using the forward direction \( \mathbf{z} = f_\theta^{-1}(\mathbf{x}) \), the density becomes:
\[
p_X(\mathbf{x}) = p_Z(\mathbf{z}) \left| \det \left( \frac{\partial f_\theta}{\partial \mathbf{z}} \right) \right|^{-1},
\]
and the corresponding log-likelihood is:
\[
\log p_X(\mathbf{x}) = \log p_Z(\mathbf{z}) - \log \left| \det \left( \frac{\partial f_\theta}{\partial \mathbf{z}} \right) \right|.
\]

In practice, a single transformation is typically not expressive enough to model complex data distributions. To address this, Normalizing Flows compose a sequence of \( K \) simple, tractable transformations:
\[
\mathbf{x} = \mathbf{z}_K = f_K \circ f_{K-1} \circ \cdots \circ f_1(\mathbf{z}_0),
\]
where each \( f_k \) is invertible and differentiable, and designed such that the Jacobian determinant is efficient to compute. The inverse computation proceeds as:
\[
\mathbf{z}_0 = f_1^{-1} \circ f_2^{-1} \circ \cdots \circ f_K^{-1}(\mathbf{x}).
\]
Given a base latent variable \( \mathbf{z}_0 \sim p_Z(\mathbf{z}) \), typically sampled from a standard multivariate Gaussian \( \mathcal{N}(0, \mathbf{I}) \), the full transformation yields the model’s output. Applying the change-of-variable formula recursively across layers, we obtain the overall log-likelihood:
\[
\log p_X(\mathbf{x}) = \log p_Z(\mathbf{z}_0) - \sum_{k=1}^K \log \left| \det \left( \frac{\partial f_k}{\partial \mathbf{z}_{k-1}} \right) \right|.
\]

\subsubsection*{Deterministic Inference and Learning}

A distinguishing feature of Normalizing Flows as PLVMs is that posterior inference is entirely deterministic. While the model defines a latent hierarchy \( \mathbf{z}_0, \dots, \mathbf{z}_K = \mathbf{x} \), each variable is a deterministic function of the others via the composed invertible mapping. Therefore, given an observation \( \mathbf{x} \), the latent variable \( \mathbf{z}_0 \) is computed exactly by inversion:
\[
\mathbf{z}_0 = f_\theta^{-1}(\mathbf{x}),
\]
yielding a posterior of the form \( p(\mathbf{z}_0 \mid \mathbf{x}) = \delta(\mathbf{z}_0 - f_\theta^{-1}(\mathbf{x})) \), where \( \delta \) is the Dirac delta function. This contrasts sharply with models like VAEs or LDA, where the posterior is intractable and must be approximated.

As a result, Normalizing Flows do not require variational inference or EM-based optimization. Instead, model parameters are learned via direct maximum likelihood estimation, typically using stochastic gradient descent. The exact log-likelihood computation provides a principled and efficient learning signal, making NFs a tractable and powerful tool for deep probabilistic modeling.

\subsection{Designing Flexible and Tractable Transformations}

While the change-of-variable formula provides a principled mechanism for exact likelihood evaluation, its practical success in Normalizing Flows hinges on the design of the component transformations \( f_k \). Each transformation must satisfy two core requirements: it must be expressive enough to model complex data distributions, and tractable enough to allow efficient computation of the Jacobian determinant required for exact likelihood evaluation.

To balance these criteria, several classes of transformations have been proposed. Two widely used examples are planar flows and affine coupling layers, which exemplify different design philosophies within Normalizing Flows.

\paragraph{Planar Flows.}

Planar flows~\cite{rezende2016variationalinferencenormalizingflows} introduce nonlinearity via a residual-style update:
\[
f(\mathbf{z}) = \mathbf{z} + \mathbf{u} \cdot h(\mathbf{w}^\top \mathbf{z} + b),
\]
where \( \mathbf{u}, \mathbf{w} \in \mathbb{R}^d \), \( b \in \mathbb{R} \), and \( h(\cdot) \) is a smooth activation function such as \( \tanh \). This transformation perturbs \( \mathbf{z} \) along a learnable direction \( \mathbf{u} \), modulated by a nonlinear projection.

The Jacobian of this transformation is:
\[
\frac{\partial f}{\partial \mathbf{z}} = \mathbf{I} + \mathbf{u} \cdot h'(\mathbf{w}^\top \mathbf{z} + b) \cdot \mathbf{w}^\top,
\]
and its log-determinant can be computed efficiently using the matrix determinant lemma:
\[
\log \left| \det \left( \frac{\partial f}{\partial \mathbf{z}} \right) \right| = \log \left| 1 + \mathbf{u}^\top \psi(\mathbf{z}) \right|, \quad \text{where } \psi(\mathbf{z}) = h'(\mathbf{w}^\top \mathbf{z} + b) \cdot \mathbf{w}.
\]

Planar flows are lightweight and analytically tractable, but due to their limited capacity per layer, they often require deep stacking to approximate complex target densities effectively.

\paragraph{Affine Coupling Layers.}

Affine coupling layers, introduced in RealNVP~\cite{dinh2017densityestimationusingreal}, take a more modular approach. The input vector is partitioned into two parts:
\[
\mathbf{z} = [\mathbf{z}_a, \mathbf{z}_b],
\]
and transformed as:
\[
\begin{aligned}
\mathbf{z}_a' &= \mathbf{z}_a, \\
\mathbf{z}_b' &= \mathbf{z}_b \odot \exp(s(\mathbf{z}_a)) + t(\mathbf{z}_a),
\end{aligned}
\]
where \( s(\cdot) \) and \( t(\cdot) \) are learned functions, typically parameterized by neural networks. Because only part of the input is transformed at a time, the Jacobian is triangular, and its log-determinant simplifies to:
\[
\log \left| \det \left( \frac{\partial f}{\partial \mathbf{z}} \right) \right| = \sum_i s_i(\mathbf{z}_a).
\]

Affine coupling layers are highly efficient, allowing for deep stacking without incurring prohibitive computational cost. The functions \( s(\cdot) \) and \( t(\cdot) \) can be implemented as simple MLPs or more sophisticated architectures such as CNNs or transformers, depending on the data modality.

Through the careful combination of these components, Normalizing Flows achieve both high expressivity and efficient, exact likelihood evaluation. These properties position them as a unique and powerful class of probabilistic latent variable models, capable of capturing complex structure in data while retaining interpretability and tractability.

\section{Diffusion Models: Sequential Probabilistic Latent Variable Models}
\label{sec:diffusion}

Diffusion models~\cite{sohl2015deep,ho2020denoisingdiffusionprobabilisticmodels} represent a distinct class of probabilistic latent variable models (PLVMs) in which the latent space is a \emph{sequence} of noisy variables bridging clean data and pure noise. The forward process \( q(\mathbf{x}_{1:T} \mid \mathbf{x}_0) \) defines a fixed, tractable Markov chain that progressively corrupts the data with Gaussian noise, while the reverse process \( p_\theta(\mathbf{x}_{0:T}) \) is a learned generative chain trained to invert this corruption.

From a PLVM perspective, the forward chain acts as a known encoder, mapping \( \mathbf{x}_0 \) to a deep latent hierarchy \( \mathbf{x}_1, \dots, \mathbf{x}_T \), and the reverse chain acts as a parameterized decoder, reconstructing \( \mathbf{x}_0 \) from \( \mathbf{x}_T \). This exact, fixed inference path differentiates diffusion models from VAEs (which learn an approximate posterior) and from NFs (which perform deterministic inversion). Their sequential latent structure and score-based training enable high-fidelity modeling of complex data distributions, underpinning recent advances in image and audio synthesis and text-to-image generation (e.g., DALL·E~\cite{marcus2022very}, Stable Diffusion~\cite{stablediffusion2024}) thanks to their ability to capture highly complex data distributions~\cite{yang2023diffusion}.

\subsection{Mechanism and Learning}

\subsection*{The Forward Diffusion Process}

The forward process defines a Markov chain of length \( T \), which gradually corrupts the observed data \( \mathbf{x}_0 \sim q(\mathbf{x}_0) \) by adding Gaussian noise at each step. Specifically, for \( t = 1, \dots, T \), the transition is defined as:

\[
q(\mathbf{x}_t \mid \mathbf{x}_{t-1}) = \mathcal{N}(\mathbf{x}_t; \sqrt{1 - \beta_t} \cdot \mathbf{x}_{t-1}, \beta_t \mathbf{I}),
\]

where \( \beta_t \in (0, 1) \) is a variance schedule controlling the noise injected at each step. As \( t \to T \), the samples become increasingly noisy, eventually approaching an isotropic Gaussian distribution.

Thanks to the linear Gaussian structure, the marginal distribution \( q(\mathbf{x}_t \mid \mathbf{x}_0) \) is also Gaussian and given in closed form by:

\[
q(\mathbf{x}_t \mid \mathbf{x}_0) = \mathcal{N}(\mathbf{x}_t; \sqrt{\bar{\alpha}_t} \cdot \mathbf{x}_0, (1 - \bar{\alpha}_t) \mathbf{I}),
\]

where the cumulative noise factor is:$
\bar{\alpha}_t = \prod_{s=1}^t (1 - \beta_s)$. This closed-form expression allows direct sampling from \( q(\mathbf{x}_t \mid \mathbf{x}_0) \) without simulating all intermediate steps.

\subsection*{The Reverse Generative Process}

The ultimate goal of diffusion models is to learn a generative process that maps Gaussian noise back into complex data, effectively reversing the forward noising process. Since the forward process is Markovian, we assume the reverse process is also Markovian and factorize the joint distribution as:

\[
p_\theta(\mathbf{x}_{0:T}) = p(\mathbf{x}_T) \prod_{t=1}^T p_\theta(\mathbf{x}_{t-1} \mid \mathbf{x}_t),
\]

where \( p(\mathbf{x}_T) = \mathcal{N}(0, \mathbf{I}) \) is the standard Gaussian prior. Each reverse transition is modeled as a Gaussian whose mean and variance are learned:

\[
p_\theta(\mathbf{x}_{t-1} \mid \mathbf{x}_t) = \mathcal{N}(\mathbf{x}_{t-1}; \bm{\mu}_\theta(\mathbf{x}_t, t), \Sigma_\theta(\mathbf{x}_t, t)).
\]

\subsection*{Learning the Reverse Process}

In principle, we aim to approximate the true posterior:

\[
q(\mathbf{x}_{t-1} \mid \mathbf{x}_t) \approx p_\theta(\mathbf{x}_{t-1} \mid \mathbf{x}_t),
\]

but this distribution is intractable due to marginalizing over all possible \( \mathbf{x}_0 \). However, by conditioning on \( \mathbf{x}_0 \), the conditional posterior \( q(\mathbf{x}_{t-1} \mid \mathbf{x}_t, \mathbf{x}_0) \) becomes analytically tractable, due to the joint Gaussian structure.

From the forward process:

\[
\mathbf{x}_t = \sqrt{1 - \beta_t} \cdot \mathbf{x}_{t-1} + \sqrt{\beta_t} \cdot \boldsymbol{\epsilon}, \quad \boldsymbol{\epsilon} \sim \mathcal{N}(0, \mathbf{I})
\]

\[
q(\mathbf{x}_{t-1} \mid \mathbf{x}_0) = \mathcal{N}(\sqrt{\bar{\alpha}_{t-1}} \cdot \mathbf{x}_0, (1 - \bar{\alpha}_{t-1}) \cdot \mathbf{I})
\]

\[
q(\mathbf{x}_t \mid \mathbf{x}_0) = \mathcal{N}(\sqrt{\bar{\alpha}_t} \cdot \mathbf{x}_0, (1 - \bar{\alpha}_t) \cdot \mathbf{I})
\]

The cross-covariance between \( \mathbf{x}_{t-1} \) and \( \mathbf{x}_t \), conditioned on \( \mathbf{x}_0 \), is:

\[
\text{Cov}[\mathbf{x}_{t-1}, \mathbf{x}_t \mid \mathbf{x}_0] = \sqrt{1 - \beta_t} \cdot (1 - \bar{\alpha}_{t-1}) \cdot \mathbf{I}.
\]

Using Gaussian conditioning identities, the posterior \( q(\mathbf{x}_{t-1} \mid \mathbf{x}_t, \mathbf{x}_0) \) is given by:

\[
q(\mathbf{x}_{t-1} \mid \mathbf{x}_t, \mathbf{x}_0) = \mathcal{N}(\mathbf{x}_{t-1}; \tilde{\bm{\mu}}_t(\mathbf{x}_t, \mathbf{x}_0), \tilde{\beta}_t \cdot \mathbf{I}),
\]

where:

\[
\tilde{\bm{\mu}}_t(\mathbf{x}_t, \mathbf{x}_0) = \frac{\sqrt{\bar{\alpha}_{t-1}} \beta_t}{1 - \bar{\alpha}_t} \cdot \mathbf{x}_0 + \frac{\sqrt{1 - \beta_t}(1 - \bar{\alpha}_{t-1})}{1 - \bar{\alpha}_t} \cdot \mathbf{x}_t,
\]

\[
\tilde{\beta}_t = \frac{1 - \bar{\alpha}_{t-1}}{1 - \bar{\alpha}_t} \cdot \beta_t.
\]

This expression forms the target distribution that the learned model \( p_\theta(\mathbf{x}_{t-1} \mid \mathbf{x}_t) \) attempts to match during training, typically by predicting \( \mathbf{x}_0 \) or the noise \( \boldsymbol{\epsilon} \) used in the forward process.

\subsection{Diffusion Models from the Lens of PLVMs}

Diffusion models can be naturally interpreted as a special class of probabilistic latent variable models (PLVMs). In the classical PLVM framework, we define a joint distribution over observed data \( \mathbf{x} \) and latent variables \( \mathbf{z} \) as:
$p(\mathbf{x}, \mathbf{z}) = p(\mathbf{z}) p_\theta(\mathbf{x} \mid \mathbf{z})$, where \( p(\mathbf{z}) \) is a prior distribution (typically a standard Gaussian), and \( p_\theta(\mathbf{x} \mid \mathbf{z}) \) is a parameterized decoder or likelihood model.

In diffusion models, the observed variable is \( \mathbf{x}_0 \) (the data), and the latent variables are the intermediate noisy representations \( \mathbf{x}_1, \dots, \mathbf{x}_T \) produced by the forward diffusion process. These latent variables form a deep structured Markov chain, rather than a single-step latent code. As introduced above, the forward noising process is defined by a fixed chain of tractable conditional Gaussians: $q(\mathbf{x}_{1:T} \mid \mathbf{x}_0) = \prod_{t=1}^T q(\mathbf{x}_t \mid \mathbf{x}_{t-1})$, and the reverse process is defined as:$   p_\theta(\mathbf{x}_{0:T}) = p(\mathbf{x}_T) \prod_{t=1}^T p_\theta(\mathbf{x}_{t-1} \mid \mathbf{x}_t)$ - this establishes a deep latent hierarchy over time, where the generation proceeds from pure noise \( \mathbf{x}_T \) toward increasingly structured intermediate states, culminating in the reconstruction of the data \( \mathbf{x}_0 \).

\subsection*{Training as Approximate Inference}

As with other PLVMs such as VAEs, the goal in diffusion models is to learn a generative model \( p_\theta(\mathbf{x}_{0:T}) \) such that the marginal distribution \( p_\theta(\mathbf{x}_0) \) approximates the true data distribution. However, a key distinction is that the inference process \( q(\mathbf{x}_{1:T} \mid \mathbf{x}_0) \) is not learned (as in amortized inference), but is fixed and fully tractable due to the Gaussian noise injection schedule.

This setting allows us to define an exact variational objective in terms of the KL divergence between the joint forward and reverse trajectories:

\[
\text{KL}(q(\mathbf{x}_{0:T}) \, \| \, p_\theta(\mathbf{x}_{0:T})) = \mathbb{E}_{q(\mathbf{x}_{0:T})} \left[ \log \frac{q(\mathbf{x}_{1:T} \mid \mathbf{x}_0)}{p_\theta(\mathbf{x}_{0:T})} \right],
\]

which corresponds to a variational lower bound (ELBO) on \( \log p_\theta(\mathbf{x}_0) \). This ELBO decomposes across timesteps into:

\[
\mathcal{L}_{\text{ELBO}} = \mathbb{E}_{q(\mathbf{x}_0)} \left[ \mathbb{E}_{q(\mathbf{x}_{1:T} \mid \mathbf{x}_0)} \left[ \sum_{t=1}^T \text{KL}\left( q(\mathbf{x}_{t-1} \mid \mathbf{x}_t, \mathbf{x}_0) \, \| \, p_\theta(\mathbf{x}_{t-1} \mid \mathbf{x}_t) \right) \right] \right].
\]

Each KL term penalizes deviation between the tractable posterior (conditioned on \( \mathbf{x}_0 \)) and the learned reverse transition. In this sense, diffusion models remain consistent with the PLVM framework: a fixed inference path and a learnable generative path optimized by minimizing divergence between conditional posteriors.

\subsection*{Score-Based Training as a Practical PLVM Objective}

While the ELBO provides a principled variational objective for training diffusion models, in practice, this formulation is often simplified for both computational and modeling efficiency. Instead of directly parameterizing the reverse transition distribution \( p_\theta(\mathbf{x}_{t-1} \mid \mathbf{x}_t) \), many diffusion models learn a neural network to approximate the underlying posterior mean by predicting a quantity related to the clean data.

Specifically, recall that the forward process defines a closed-form conditional:

\[
\mathbf{x}_t = \sqrt{\bar{\alpha}_t} \cdot \mathbf{x}_0 + \sqrt{1 - \bar{\alpha}_t} \cdot \boldsymbol{\epsilon}, \quad \boldsymbol{\epsilon} \sim \mathcal{N}(0, \mathbf{I}),
\]

This formulation allows us to reparameterize training as a denoising task. Instead of regressing toward the exact posterior mean \( \tilde{\bm{\mu}}_t(\mathbf{x}_t, \mathbf{x}_0) \), we train a neural network \( \epsilon_\theta(\mathbf{x}_t, t) \) to predict the noise \( \boldsymbol{\epsilon} \) that was used to generate \( \mathbf{x}_t \) from \( \mathbf{x}_0 \). This leads to the simplified denoising score-matching loss:

\[
\mathcal{L}_{\text{simple}} = \mathbb{E}_{\mathbf{x}_0, \boldsymbol{\epsilon}, t} \left[ \left\| \boldsymbol{\epsilon} - \epsilon_\theta(\mathbf{x}_t, t) \right\|^2 \right],
\]

This objective is equivalent to minimizing a particular weighting of the full ELBO, where all Gaussian variances are fixed to constant values. It can also be interpreted as a form of score matching: instead of minimizing KL divergence between posteriors, the model is trained to match the gradient of the log-density \( \nabla_{\mathbf{x}_t} \log q(\mathbf{x}_t \mid \mathbf{x}_0) \), which corresponds to the optimal denoising direction under the Gaussian corruption.

From the PLVM perspective, this approach retains the same structure in that 1) The forward path \( q(\mathbf{x}_{1:T} \mid \mathbf{x}_0) \) acts as an implicit encoder. 2) The model learns to reconstruct or denoise \( \mathbf{x}_0 \) from \( \mathbf{x}_t \) — either directly or via the noise \( \boldsymbol{\epsilon} \). 3) The training signal is derived from the known posterior \( q(\mathbf{x}_{t-1} \mid \mathbf{x}_t, \mathbf{x}_0) \), which remains analytically available.

As derived earlier, this posterior is Gaussian:
$q(\mathbf{x}_{t-1} \mid \mathbf{x}_t, \mathbf{x}_0) = \mathcal{N}(\tilde{\bm{\mu}}_t(\mathbf{x}_t, \mathbf{x}_0), \tilde{\beta}_t \mathbf{I})$, where \( \tilde{\bm{\mu}}_t \) is a known combination of \( \mathbf{x}_0 \) and \( \mathbf{x}_t \), determined by the noise schedule. Thus, learning to predict either \( \mathbf{x}_0 \) or \( \boldsymbol{\epsilon} \) from \( \mathbf{x}_t \) implicitly reconstructs the posterior mean, enabling supervised training.

In summary, score-based training in diffusion models emerges as a tractable, supervision-friendly simplification of the full PLVM-style ELBO, where the variational inference is effectively amortized via a denoising network, and the encoder is embedded in the known forward process.

Diffusion models can thus be interpreted as PLVMs with: 1) A deep latent chain of variables \( \mathbf{x}_1, \dots, \mathbf{x}_T \) representing noisy versions of the data; 2) A tractable and fixed inference process \( q \) defined by Gaussian corruption; 3) A parameterized generative model \( p_\theta \) trained to reverse this corruption. This view reveals strong connections to classical probabilistic modeling frameworks, while also highlighting how diffusion models uniquely leverage tractable noising and sampling dynamics for flexible, high-fidelity generative modeling.

\section{Autoregressive Models: Explicit Probabilistic Generative Models}\label{sec:autoregressive}

Autoregressive (AR) models represent one of the most fundamental and widely used classes of generative models, grounded in the chain rule of probability. Rather than relying on latent variables or invertible mappings, autoregressive models directly model the joint distribution of high-dimensional data as an explicit product of conditionals over individual dimensions or time steps. For a data point \( \mathbf{x} = (x_1, x_2, \dots, x_D) \), the joint density is factorized as:$
p_\theta(\mathbf{x}) = \prod_{d=1}^D p_\theta(x_d \mid x_{<d})$,
where \( x_{<d} \) denotes all preceding variables in some fixed ordering. Each conditional \( p_\theta(x_d \mid x_{<d}) \) is parameterized by a neural network, allowing the model to capture complex, multimodal, and high-dimensional distributions with exact likelihood computation.

Autoregressive models differ fundamentally from VAEs, Normalizing Flows, and diffusion models in how they approach inference and likelihood computation.
From the PLVM perspective, AR models define a fully observable generative process. The sequential factorization can be interpreted as an implicit latent trajectory over partially revealed variables, but no unobserved random variables are introduced. 
This formulation avoids the need for approximate inference entirely and enables exact maximum likelihood training, via a product of conditionals, enabling exact and tractable log-likelihood computation: $\log p_\theta(\mathbf{x}) = \sum_{d=1}^D \log p_\theta(x_d \mid x_{<d})$. This decomposition allows the model to be trained using supervised learning techniques, where each conditional is treated as a prediction task given the context of preceding variables. This approach—commonly referred to as teacher forcing—enables efficient and stable training, since the true preceding values \( x_{<d} \) are always known during training and can be fed into the model in parallel.

In contrast, Variational Autoencoders (VAEs) introduce latent variables to model global structure, but at the cost of intractable posteriors \( p(\mathbf{z} \mid \mathbf{x}) \), which must be approximated via amortized variational inference. %Although VAEs are efficient to train and sample from, their reliance on approximate posteriors can limit modeling fidelity, especially in high-precision domains like image generation.
Diffusion models, meanwhile, model the data as the endpoint of a gradual denoising process and optimize a variational lower bound through score matching. Although diffusion models achieve state-of-the-art generation quality across many domains, their inference is inherently iterative and computationally intensive, requiring dozens to hundreds of sequential denoising steps to generate a sample. %Exact likelihood computation is possible but more complex and often de-emphasized in favor of high-quality synthesis.
Normalizing Flows offer an exact alternative by constructing invertible transformations between latent and observed variables. This invertibility enables exact posterior inference and likelihood computation, but imposes architectural constraints on the transformations to maintain tractable Jacobians. %While flows are more expressive than VAEs under exact training, their sampling and training remain less parallelizable than autoregressive models in practice.

Architecturally, AR models have been implemented using recurrent networks (e.g., RNN language models), masked convolutions (e.g., PixelCNN~\cite{oord2016conditionalimagegenerationpixelcnn} for images, WaveNet~\cite{oord2016wavenetgenerativemodelraw} for audio), and more recently, self-attention mechanisms (e.g., the GPT series~\cite{openai2024gpt4technicalreport} for text). These variants differ in how they parameterize \( p_\theta(x_d \mid x_{<d}) \) and in the trade-off between parallelism during training and the expressiveness of long-range dependency modeling. Overall, AR models offer tractability in both training and likelihood evaluation, making them especially well-suited for tasks requiring precise density estimation or modeling of complex sequential dependencies. By factorizing the joint distribution into a product of conditionals, they allow exact log-likelihood computation and stable, straightforward training. Models such as PixelCNN and the GPT series have leveraged this structure to achieve state-of-the-art performance across diverse domains including image generation, audio synthesis, and natural language processing.

A principal limitation of AR models lies in their sequential sampling procedure: generating a new data point requires sampling each dimension or timestep one at a time, which becomes a computational bottleneck in high-dimensional domains such as images or raw audio. This has motivated research into parallel decoding schemes, distillation-based speedups, and hybrid architectures that integrate autoregressive components with latent-variable models (e.g., VAE-AR hybrids~\cite{cai2023hybrid}) to balance tractability, sampling efficiency, and representational power.

\section{Generative Adversarial Networks: Implicit Probabilistic Latent Variable Models}\label{sec:GANs}

Generative Adversarial Networks (GANs)~\cite{goodfellow2014generative} represent a fundamentally different approach to generative modeling—eschewing explicit likelihoods in favor of adversarial learning. Rather than defining or maximizing a tractable probability density, GANs train a generator to produce samples indistinguishable from real data according to an adaptive discriminator. This framing makes GANs an example of \emph{implicit probabilistic latent variable models}: the model introduces latent variables but does not provide a closed-form likelihood or posterior.

In the standard formulation, a latent variable \(\mathbf{z} \sim p_Z(\mathbf{z})\) (often Gaussian or uniform) is mapped deterministically to the data space by a generator \(G_\theta(\mathbf{z})\). A discriminator \(D_\phi(\mathbf{x})\) learns to distinguish real samples from generated ones. The networks are trained in a two-player minimax game:
\[
\min_\theta \max_\phi \; \mathbb{E}_{\mathbf{x} \sim p_{\text{data}}}[\log D_\phi(\mathbf{x})] 
+ \mathbb{E}_{\mathbf{z} \sim p_Z}[\log(1 - D_\phi(G_\theta(\mathbf{z})))],
\]
where the generator improves by fooling the discriminator, and the discriminator improves by correctly detecting synthetic data. No explicit density \(p_\theta(\mathbf{x})\) is computed; instead, learning proceeds entirely through sample-based feedback.

From the PLVM perspective, GANs can be interpreted as models with latent variables \( \mathbf{z} \sim p_Z(\mathbf{z}) \) and a deterministic decoder \( \mathbf{x} = G_\theta(\mathbf{z}) \), but without an explicitly defined or tractable likelihood \( p_\theta(\mathbf{x}) \). In this sense, they correspond to \textit{implicit latent variable models}, where inference is bypassed entirely and learning proceeds via sample-level feedback rather than likelihood maximization.

GANs excel in producing visually sharp, high-fidelity samples, often surpassing likelihood-based models in perceptual quality, and have driven advances in photo-realistic image synthesis~\cite{karras2019stylebasedgeneratorarchitecturegenerative}, image-to-image translation~\cite{gong2024e2ganefficienttrainingefficient}, and super-resolution~\cite{tian2024generativeadversarialnetworksimage}. Nevertheless, the absence of a likelihood objective presents notable challenges: evaluation must be performed indirectly, relying on surrogate metrics such as the Inception Score (IS) or Fréchet Inception Distance (FID) rather than exact log-likelihoods, and training stability can be fragile, with issues such as mode collapse, vanishing gradients, and non-convergent adversarial dynamics frequently encountered. Numerous refinements have emerged to address these limitations, such as Wasserstein GANs~\cite{arjovsky2017wassersteingan}, StyleGAN~\cite{karras2019stylebasedgeneratorarchitecturegenerative}, and hybrids combining GANs with diffusion~\cite{wang2023diffusiongantraininggansdiffusion} or autoregressive components~\cite{eskandarinasab2024seriesgantimeseriesgeneration}. In practice, GANs are particularly well-suited to domains where sample quality and diversity take precedence over tractable likelihood estimation or explicit inference.

\section{Conclusion}
\label{sec:conclusion}

This paper set out to provide a conceptual roadmap for generative AI through the unifying lens of probabilistic latent variable models (PLVMs). By tracing the progression from classical flat models, such as probabilistic PCA, Gaussian mixture models, latent class analysis, item response theory, and latent Dirichlet allocation, to their sequential extensions, including Hidden Markov Models, Gaussian HMMs, and Linear Dynamical Systems, and finally to modern deep architectures, we have shown how core latent-variable principles—density estimation, latent reasoning, and structured inference—remain central to state-of-the-art methods.

Within this taxonomy, Variational Autoencoders were interpreted as Deep PLVMs, where neural parameterizations extend the flexibility of classical formulations but require approximate inference. Normalizing Flows emerged as Tractable PLVMs, offering exact likelihood computation and deterministic inference through invertible mappings. Diffusion Models were cast as Sequential PLVMs, generating data via stepwise latent evolution optimized with score matching. Autoregressive Models represented Explicit Probabilistic Generative Models, factorizing the joint distribution into fully observable conditionals. Finally, Generative Adversarial Networks were positioned as Implicit PLVMs, forgoing explicit likelihoods in favor of adversarial training on sample quality.

Framing these models within a single probabilistic scaffold reveals both their shared foundations and their key differences in inference strategy, representational capacity, and computational trade-offs. This synthesis provides a conceptual roadmap for understanding existing methods, guiding model selection, and inspiring the design of novel hybrids.

By situating modern architectures within their probabilistic lineage, including both flat and sequential classical models, we hope this work serves as both a coherent theoretical framework and a springboard for methodological innovation. The continued interplay between probabilistic modeling and deep learning promises to shape the next generation of generative AI systems—models that are not only more capable, but also more interpretable, efficient, and aligned with human values.

\end{document}